\documentclass[10pt]{article}

\usepackage[a4paper,margin=1in]{geometry}

\usepackage[T1]{fontenc}
\usepackage[utf8]{inputenc}
\usepackage{lmodern}
\usepackage{microtype}

\usepackage{amsmath,amssymb,amsthm}
\usepackage{algorithm}
\usepackage{algpseudocode}

\usepackage{graphicx}
\usepackage{booktabs}
\usepackage{array}
\usepackage{adjustbox}
\usepackage{tabularx}
\usepackage{longtable}
\usepackage{pdflscape}
\usepackage{tikz}
\usepackage{caption}
\usepackage{subcaption}
\usepackage{ragged2e}
\usepackage{multicol}
\usepackage{xurl}
\usepackage{makecell}
\usepackage{threeparttable}
\usepackage{multirow}
\usepackage[section]{placeins}
\usepackage{comment}
\usepackage{cuted}
\usepackage{xcolor}

\usepackage[numbers]{natbib}
\usepackage{hyperref}

\usetikzlibrary{arrows.meta,positioning,shapes.geometric}
\graphicspath{{figures/}}

\hypersetup{
  breaklinks=true,
  colorlinks=true,
  linkcolor=black,
  citecolor=black,
  filecolor=black,
  urlcolor=black
}

\newcolumntype{L}[1]{>{\RaggedRight\arraybackslash}p{#1}}
\newcolumntype{Y}{>{\RaggedRight\arraybackslash}X}
\newcolumntype{C}[1]{>{\centering\arraybackslash}p{#1}}

\newcommand{\nomgroup}[1]{\medskip\noindent\textbf{#1}\par\smallskip}
\newcommand{\nomentry}[2]{%
  \noindent
  \begin{tabularx}{\linewidth}{@{}>{\RaggedRight\arraybackslash}p{0.2\linewidth}>{\RaggedRight\arraybackslash}X@{}}
  #1 & #2
  \end{tabularx}
  \par\vspace{0.15em}
}

\newenvironment{keywords}
{\vspace{0.5em}\noindent\textbf{Keywords:}\ }
{\par\vspace{0.5em}}

\title{\Large When Agents Meet Electric Bus Fleet Operations: Pricing Behavior, Trade-offs, and Policy Implications in an Aggregator Framework}

\author{\small
J\^onatas Augusto Manzolli$^{1,2,*}$ \quad
Ali Eslami$^{1,*}$ \quad
Luis Miranda-Moreno$^{1}$ \quad
Jiangbo Yu$^{1,\dagger}$\\[0.6em]
\footnotesize $^{1}$Department of Civil Engineering, McGill University, Montreal, Quebec, Canada\\
\footnotesize $^{2}$INESC Coimbra, University of Coimbra, Coimbra, Portugal\\[0.4em]
\footnotesize $^{*}$These authors contributed equally to this work. \quad
\footnotesize $^{\dagger}$Corresponding author: \texttt{jiangbo.yu@mcgill.ca}
}

\date{}

\begin{document}

\maketitle

\begin{abstract}
Agentic systems are changing how complex operational tasks are coordinated by connecting heterogeneous data sources, introducing a new paradigm for simplifying tasks, connecting datasets, and automating processes. Electric bus fleets provide a relevant test case for this paradigm. Their operation requires continuous coordination between service reliability, battery state-of-charge, charger availability, electricity prices, route-energy uncertainty, and vehicle-to-grid (V2G) opportunities. This paper proposes an agentic aggregator framework that streamlines this decision environment by coupling an optimization-based electric bus scheduling model with supervisory agents for disturbance detection, tariff adaptation, and schedule evaluation. The optimization core enforces physical feasibility across routes, chargers, batteries, and V2G exchanges, while the agentic layer interprets changing operating conditions, triggers real-time re-optimization when needed, and defines how flexibility value is allocated between the aggregator and the public transport operator (PTO). A realistic depot case study evaluates day-ahead and real-time operations under profit-based and operation-based coordination modes, considering service delays, route-energy deviations, electricity price shocks, and combined disturbances. The results show that agentic aggregation can support adaptive fleet-grid coordination by maintaining feasible schedules, activating re-optimization selectively, and improving the use of charging and V2G flexibility. However, they also reveal a critical trade-off: the same agentic capability that reduces operational complexity can extract value from the PTO when configured around profit-oriented pricing. Prompt-sensitivity experiments further show that agentic pricing behavior changes with prompt configuration, making prompt design a reproducibility and governance variable. These findings suggest that agentic aggregators can become useful tools for managing electric bus V2G operations, but their deployment in public-fleet contexts requires transparent coordination modes, auditable tariff-setting, and explicit value-sharing rules.
\end{abstract}

\begin{keywords}
Electric bus fleets; Aggregators; Smart charging; Vehicle-to-grid; Energy market; Agentic AI
\end{keywords}


\section{Introduction}
\label{sec:introduction}

Artificial intelligence (AI) is rapidly moving from passive decision support toward agentic systems capable of coordinating tasks, connecting heterogeneous data sources, invoking external tools, and automating structured workflows. This evolution is particularly relevant for energy and transport systems, where operational decisions depend on fragmented information, physical constraints, market signals, and real-time disturbances. In such contexts, agentic AI should not be understood as a replacement for mathematical optimization or engineering models, but as a supervisory layer that can streamline the complex workflow around them: detecting when conditions change, preparing decision inputs, selecting the appropriate analytical tool, and interpreting the resulting outputs.

Electric bus fleets provide a timely test case for this capability. As public transport electrification expands, electric buses are becoming large, time-varying loads connected to the power system \cite{iea2024global}. Their operation is already complex because public transport operators (PTOs) must coordinate battery state-of-charge (SOC), charger availability, route assignments, timetable reliability, charging duration, route-energy uncertainty, and depot power limits \cite{manzolliReviewElectricBus2022}. This complexity increases further when vehicle-to-grid (V2G) operation and energy trading are introduced. The fleet must then decide not only when to charge, but also when to discharge, how much energy to export, and under which tariff or compensation structure participation remains attractive. Previous optimization-based studies have shown that coordinated charging, tariff-aware scheduling, and V2G strategies can reduce costs and improve system performance, especially when battery aging, demand peaks, and uncertainty are explicitly considered \cite{manzolli2025robust,sadeghian2022comprehensive}. However, deploying these strategies in practice remains challenging because the required information changes continuously during operation.

In this setting, the aggregator becomes a natural coordination entity \cite{soaresRoleAggregatorsEnergy2022}. In electricity systems, aggregators pool distributed flexibility, interact with markets, and translate system-level signals into coordinated decisions across users and assets \cite{burger2017value}. For electric bus fleets, the aggregator can be understood as an intermediary between grid needs and fleet operations. It receives market and grid signals, converts them into fleet-feasible charging and V2G actions, and defines how the economic value of flexibility is shared between the aggregator and the PTO. This role is broader than simple arbitrage because the value of flexibility cannot be separated from service reliability, terminal SOC, charger access, battery feasibility, and the willingness of the PTO to participate.

Existing studies have modeled aggregator--fleet interaction through hierarchical or market-oriented formulations, including leader--follower structures in which an aggregator defines trading conditions and the fleet operator responds with charging and discharging decisions \cite{bruninxInteractionAggregatorsElectricity2020, amamraVehicletoGridAggregatorSupport2019, manzolliElectricBusSmart2022}. These models provide an important mathematical basis for aggregator-supported electric bus operation. However, they are usually built around pre-structured inputs, fixed decision horizons, and predefined optimization calls. As a result, they are useful for planning and scenario analysis, but less suited to real-time environments where delays, route-energy deviations, vehicle-state uncertainty, and electricity-price changes can rapidly make a previous schedule outdated. Existing electric bus charging studies offer powerful formulations for cost minimization, infrastructure use, and fleet scheduling, but they place less emphasis on the orchestration layer that decides when and how these models should be used in operation \cite{tianIntegratedAnalysisModeling2025a,zhaoLargescaleElectricBus2025}.

Agentic systems can address this gap by being embedded within the aggregator. Instead of treating the aggregator only as a tariff-setting optimization problem, an agentic aggregator can operate as a structured supervisory system around the charging optimizer. A Trigger Agent can monitor operational and market deviations and decide whether re-optimization is needed. A Pricing Agent can translate the selected coordination strategy into buy and sell multipliers for charging and V2G exchange. An Evaluator Agent can assess whether the resulting schedule is acceptable from economic and operational perspectives. In this architecture, the optimizer remains responsible for enforcing feasibility, while the agentic layer simplifies the surrounding decision workflow. The potential benefit is operational and economic: the system can coordinate charging and V2G events more adaptively, reduce manual intervention, and allocate flexibility value between the aggregator and the PTO.

At the same time, this capability creates a critical concern. If an agentic aggregator controls pricing guidance, its behavior determines who captures the value of V2G flexibility. The same system that can streamline electric bus operation and protect the PTO can also be configured to extract value from it through profit-oriented tariff behavior. This concern is particularly important because agentic behavior may depend on prompt configuration and coordination-mode instructions, which are not necessarily visible to the PTO or easy for regulators to audit. Therefore, the central question is not only whether agentic systems can operate electric bus V2G, but also how their behavior affects coordination, pricing, and value allocation.

Despite the rapid emergence of agentic AI in energy and infrastructure applications \cite{yu2025preparing,yu2025agentic,eslami_control-theoretic_2026,manzolliSyntheticMulticriteriaDecision2025}, the literature still lacks a concrete evaluation of an agentic aggregator for electric bus fleet operation. In particular, there is limited evidence on how such a system should be structured, how it should interact with an optimization-based charging model, how it behaves under day-ahead (DA) and real-time (RT) operation, and how different pricing behaviors affect both the aggregator and the PTO. There is also little quantitative evidence on whether agentic pricing configurations can shift value from public fleet operators to aggregators under otherwise feasible operating schedules.

Motivated by this gap, this paper proposes an agentic aggregator framework for electric bus charging and V2G coordination. The framework combines an optimization-based PTO scheduling model with three supervisory agents for triggering, pricing, and schedule evaluation. The system is tested under DA and RT conditions, including service-timing disturbances, route-energy deviations, electricity-price shocks, and combined stress cases. Two coordination modes are compared: a profit-based mode that prioritizes aggregator revenue and an operational-based mode that prioritizes PTO-compatible flexibility participation. The main contributions of this paper are:

\begin{itemize}
	\item A traceable agentic-AI architecture for adaptive EV aggregation that combines optimization with Trigger, Pricing, and Evaluator Agents into a unified pipeline separating disturbance detection, tariff 
	adaptation, and schedule acceptance.
	\item A controlled DA and RT evaluation comparing profit-based and operational-based aggregator coordination modes, showing how agentic pricing reallocates flexibility value between the aggregator and the PTO across delay, energy-consumption, price, and combined disturbances.
	\item An applied assessment of agentic aggregator pricing schemes and prompt configurations, deriving policy and market-design implications for tariff transparency, value-sharing, and PTO-compatible participation in V2G flexibility markets.
\end{itemize}

The remainder of this paper is organized as follows. Section~\ref{sec:review} reviews the relevant literature on electric bus charging, aggregation mechanisms, and emerging agentic AI applications in energy systems. Section~\ref{sec:framework} presents the proposed methodology, including the framework architecture, agentic supervisory logic, and optimization interface. Section~\ref{sec:case} describes the case study used to instantiate the framework. Section~\ref{sec:results} reports and discusses the main results, including DA operation, RT disturbances, pricing behavior, and prompt sensitivity. Section~\ref{sec:conclusion} concludes the paper, discusses policy implications, and outlines directions for future research.

\section{Literature Review}
\label{sec:review}

The transition toward electric bus systems represents a fundamental shift in public transport, extending beyond vehicle electrification to the coordinated management of fleets, infrastructure, and energy systems. As electrification scales, operational decisions become increasingly coupled with electricity markets, grid constraints, and multi-actor interactions, requiring integrated approaches that combine optimization, economic coordination, and system-level intelligence. This literature review examines three converging research streams: electric bus optimization-based charging and pricing strategies, fleet–grid interaction through aggregation mechanisms, and emerging agentic AI frameworks for system orchestration.

\subsection{Optimization-Based Charging and Pricing Strategies}

Optimization remains the dominant analytical paradigm for electric bus charging management. A large part of the literature focuses on scheduling charging activities under battery constraints, charging windows, electricity tariffs, and infrastructure limits. Representative examples include integrated models for charger deployment and fleet scheduling \cite{wang2022integrated}, general charging scheduling formulations under time-of-use tariffs and station capacity constraints \cite{xie2023optimalscheduling}, and facility planning approaches that explicitly consider uncertainty in travel times, battery degradation, and charger technology choices \cite{zhou2023facilityplanning}. Other studies extend the operational scope by incorporating seasonality, power matching, and infrastructure design \cite{liu2021seasonality}, by jointly optimizing fleet composition and scheduling across multiple depots and charging technologies \cite{yildirim2021fleet}, or by handling large network-scale charging schedules through branch-and-price and adaptive neighborhood search methods \cite{zhou2024network}. Recent work also shows growing interest in scalable heuristics for real-world agencies, with applications to fleet electrification and charger allocation under time-varying electricity prices \cite{naeimian2025decomposition}. Another important stream emphasizes operational realism beyond the minimization of daily charging costs. Battery degradation has been incorporated through both dynamic programming approaches that match workloads to battery aging \cite{jing2020capacityfade} and formulations that embed degradation directly into network-scale scheduling \cite{manzolliOptimisationElectricBus2022}. Related studies also explore coupling bus charging with local energy resources or grid services. For example, battery charging and discharging, demand response, and renewable integration have been examined in public transport settings \cite{ke2020demandresponse}, while photovoltaic-storage-charging coordination has been proposed to reduce external grid purchases and improve local energy autonomy \cite{hu2025psc}.

\subsection{Electric Bus Aggregation and Fleet--Grid Interaction}

Lately, research on electric bus systems has increasingly shifted from vehicle-level feasibility questions to fleet-level coordination with charging infrastructure and electricity markets. Early studies established that the operational viability of electric buses depends strongly on route characteristics, charging opportunities, infrastructure siting, and realistic estimates of energy consumption \cite{pagliaro2019electric,gallet2018energy}. More recent review work confirms that electric bus scheduling has evolved into a multi-layer problem involving vehicle assignment, charging management, infrastructure design, and robustness considerations rather than a simple charging-timing exercise \cite{zhang2023reviewebus}. As the scale of electrification increases, these issues become inseparable from grid interaction, since concentrated depot charging and opportunity charging can create large, time-varying loads with direct implications for peak demand, flexibility provision, and local network planning \cite{faiaMultiagentBasedEnergy2023}. In this broader context, the aggregator serves as a useful intermediary, consolidating flexibility, translating market signals, and coordinating assets that would otherwise be difficult for a PTO to manage directly \cite{carreiro2017aggregators}. For example, Cao et al. \cite{caoOptimalSchedulingElectric2020} developed a robust optimization model for scheduling an EV aggregator under upstream market price uncertainty, showing that different charging and discharging strategies can protect the aggregator's profits across optimistic, deterministic, and pessimistic price scenarios. Clairand et al. \cite{clairand2022busaggregator} showed that accounting for the aggregator in charging station planning can reduce energy costs while respecting grid constraints. Chen and Strunz \cite{chenOptimalElectricBus2025} proposed an aggregator-based framework for coordinating multi-area, multifunctional electric bus charging stations that support normal charging, fast charging, and battery swapping, while integrating renewable energy procurement and frequency-control ancillary services to reduce operating costs. More recently, the framework of Manzolli et al. \cite{manzolli2024hierarchical} explicitly modeled the interaction between an aggregator and a PTO through a hierarchical optimization structure. In that formulation, the aggregator participates in energy trading and defines pricing signals, while the PTO schedules charging and discharging activities subject to service requirements. Taken together, these studies support the idea that fleet-grid interaction is not only a charging problem, but also an intermediation and coordination problem in which the aggregator plays a central role. However, these studies usually assume that the relevant data are already available and that the decision to rerun the optimization is externally determined rather than reasoned about within the framework itself. Moreover, these studies treat the aggregator's pricing role as a technical design choice rather than a source of economic risk to the PTO, leaving the regulatory implications of aggregator tariff-setting power largely unexamined.

\subsection{Agentic Orchestration for Energy Systems}
Recent advances in agentic AI, enabled largely by large language models (LLMs), have opened a new line of inquiry into their roles in energy and cyber-physical systems. Existing studies generally do not argue that these models should replace formal optimization or physics-based models. Instead, they suggest that they may be useful as semantic and decision-support layers that can interpret heterogeneous information, support operator interaction, orchestrate tools, and improve monitoring workflows. Majumder et al. \cite{majumder2024llmenergy} discuss both the promise and the limitations of LLMs in the electric energy sector, emphasizing their usefulness for knowledge-intensive tasks while also warning about hallucinations, reliability, and domain grounding. Zhang et al. \cite{zhang2026llmenergy} similarly frame language-enabled AI as an enabling technology for information integration, decision support, and human--AI collaboration across energy applications. Emerging review work on smart grids also highlights the relevance of agentic AI to grid operations, market analysis, security support, and adaptive energy services when deployed within carefully designed supervisory architectures \cite{shi2024smartgrids,antonesi2025agentic}. Although this literature is still young, a common message is already emerging: the comparative advantage of agentic AI lies less in directly solving constrained optimization problems and more in orchestrating complex information flows around specialized analytical tools. This is especially relevant for energy systems where data are fragmented across operational databases, market signals, textual procedures, alarms, and engineering models. For EV aggregation, such capabilities could support state 
interpretation, pricing-policy enforcement, and the triggering of 
optimization runs under changing system conditions. However, concrete frameworks that connect multi-agent supervisory reasoning to electric bus fleet optimization remain scarce in the peer-reviewed literature.

\subsection{Research Gap}


The reviewed literature reveals five main observations. First, studies on electric buses have already demonstrated the importance of fleet-aware charging coordination, fleet planning, and realistic operational constraints. Second, aggregator-oriented research has shown that intermediary entities can improve the economic and operational integration of flexible loads with electricity markets. Third, related optimization work has advanced considerably in areas such as uncertainty modeling, infrastructure design, battery degradation, and market-responsive scheduling. Fourth, the emerging AI-for-energy literature suggests that agentic supervisory systems may be useful for orchestration, interpretation, and decision support, but it remains mostly conceptual or broadly scoped at the grid level. Fifth, neither the aggregator-fleet literature nor the emerging agentic AI literature has examined the regulatory implications of deploying AI-driven pricing agents in energy markets, systems whose pricing aggressiveness can be reconfigured through prompt design alone, without any structural change visible to regulators or PTOs.

What remains is a framework that integrates these strands into a unified operational architecture. In particular, the literature still lacks an approach that simultaneously integrates: i) aggregator-fleet optimization grounded in electric bus operational constraints; ii) context-aware orchestration of heterogeneous fleet and market information; iii) event-triggered re-optimization based on deviations, disturbances, or opportunity signals; and iv) coordination-mode reasoning in the pricing layer, such as contrasting operational-based and profit-based aggregator strategies. The literature also lacks any quantitative assessment of how aggregator coordination mode and prompt configuration affect PTO economic exposure, and what regulatory measures this implies. This paper addresses these gaps by embedding a multi-agent supervisory layer into the aggregator setting, while preserving the optimization engine as the constraint-enforcing core of the decision process, and by drawing explicit policy implications from the quantitative comparison of aggregator coordination modes.



\section{Methodology}
\label{sec:framework}

The proposed methodology is organized around the separation between
agentic supervision and optimization-based feasibility enforcement.
Section~\ref{sec:overallframework} introduces the overall framework
architecture and explains how the fleet, grid, and agentic aggregator
layer interact. Section~\ref{sec:multiagent} details the agentic
aggregator framework, including the supervisory agents, coordination
modes, and event-triggered re-optimization logic.
Section~\ref{sec:pto_summary} then presents the compact PTO scheduling
and optimization interface governed by the agentic layer. ~\ref{sec:bilevel} provides the full optimization formulation,
the real-time re-optimization updates, and the formal supervisory flow
that connects the agentic layer to the optimizer.

\subsection{Overall Framework}
\label{sec:overallframework}

\begin{figure*}[t]
    \centering
    \includegraphics[width=0.95\textwidth]{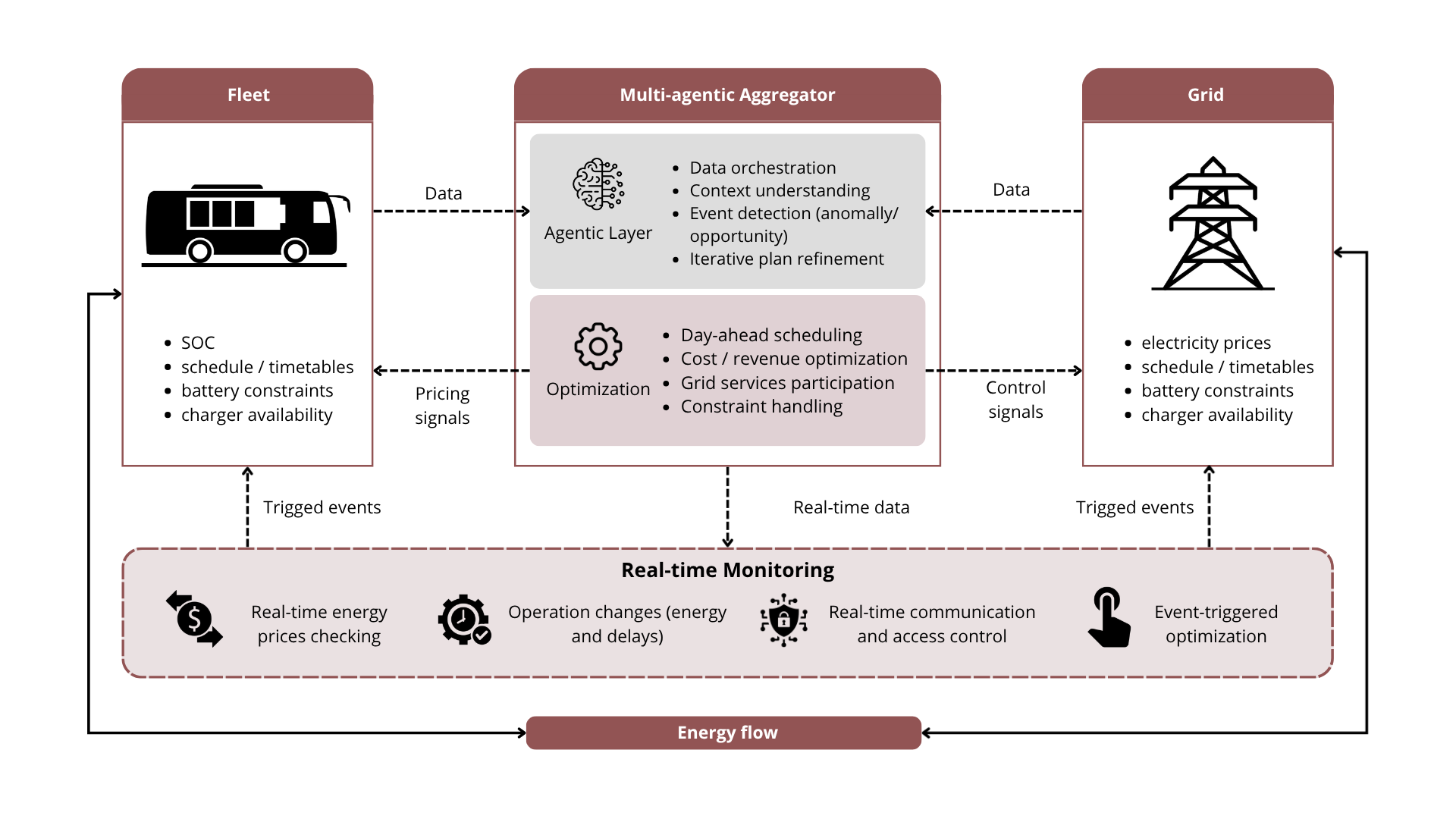}
    \caption{Proposed framework for electric bus fleet--grid
    interaction coordinated by the agentic aggregator.}
    \label{fig:framework}
\end{figure*}

As illustrated in Figure~\ref{fig:framework}, the proposed framework is
organized around three coupled components: the fleet and charging
system, the grid and market environment, and the multi-agentic
aggregator layer that coordinates the interaction between them.

On the fleet side, the framework receives operational information,
including bus state-of-charge (SOC), battery limits, charger
availability, trip assignments, departure and arrival times, and
disturbance information related to timetable delays and route-energy
deviations. These data define the physical feasibility region of the
problem, since charging and discharging decisions must preserve
transport service while respecting battery and infrastructure
constraints. On the grid and market side, the framework receives
electricity prices, power limits, and V2G flexibility opportunities.
These two information streams differ in format, timescale, and
reliability: fleet states may change during service execution, whereas
market signals may vary at hourly or sub-hourly intervals.

The aggregator layer acts as the coordination interface between the
fleet and the grid. In the day-ahead (DA) workflow, the Pricing Agent
translates the selected coordination mode into structured pricing
guidance, and the Evaluator Agent assesses the resulting schedule
against economic and operational criteria. In the real-time (RT)
workflow, the Trigger Agent monitors deviations from the DA reference
and decides whether operating conditions justify re-optimization. If
re-optimization is triggered, the Pricing Agent and Evaluator Agent are
invoked again over the remaining horizon.

This architecture preserves the main strengths of optimization-based
fleet scheduling, namely explicit constraints, reproducible decisions,
and feasibility guarantees, while adding a contextual reasoning layer
for event-driven operation. The optimization engine remains the
decision core for computing feasible charging and discharging schedules.
The multi-agentic layer does not replace this optimizer; rather, it
acts as a supervisory mechanism that determines when the optimizer
should be called, what pricing posture should be used, and whether the
resulting schedule should be accepted.

\subsection{Agentic Aggregator Framework}
\label{sec:multiagent}

In the proposed framework, the aggregator's pricing and coordination
function is performed by specialized supervisory agents rather than by
a direct optimization over tariff variables. The agents produce
structured decisions around a constrained PTO scheduling interface,
which is summarized in Section~\ref{sec:pto_summary} and fully stated
in \ref{sec:bilevel}. This separation is important: the
agents govern pricing guidance, triggering, and schedule acceptance,
whereas the optimizer enforces the operational constraints of the bus
fleet, chargers, batteries, and V2G exchanges.

At the agentic level, the price-guidance vector $\mathbf{u}_{\tau}$
contains bounded buy and sell multipliers for each tariff period. Given
this guidance and the current fleet--grid context, the optimizer
returns a candidate schedule $\mathbf{s}_{\tau}^{*}$ and the associated
KPIs, including PTO cost, aggregator revenue, V2G activity, feasibility
status, and terminal SOC. The tariff mapping and the compact PTO
optimization model are introduced in Section~\ref{sec:pto_summary}.

\begin{figure*}[t]
    \centering
    \begin{subfigure}[t]{0.75\textwidth}
        \centering
        \includegraphics[width=\textwidth,trim=12 8 12 8,clip]{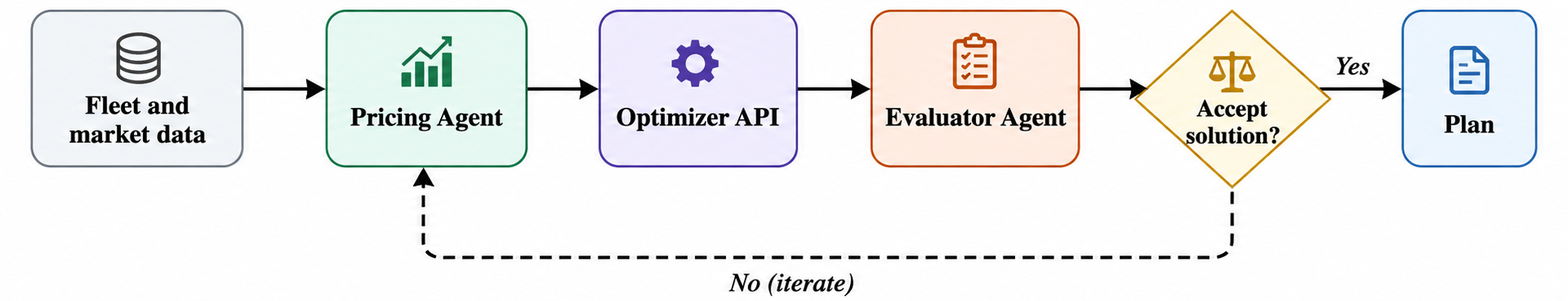}
        \caption{DA architecture.}
    \end{subfigure}

    \vspace{0.35em}

    \begin{subfigure}[t]{0.75\textwidth}
        \centering
        \includegraphics[width=\textwidth,trim=12 8 12 8,clip]{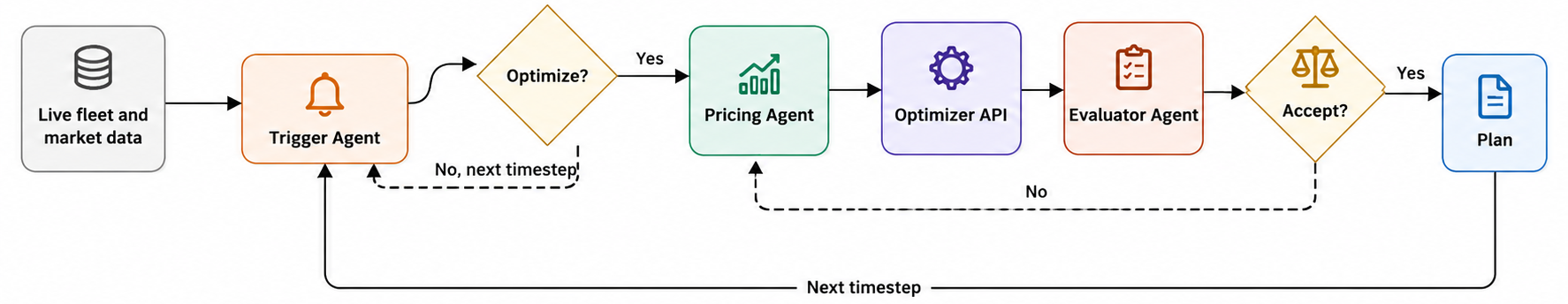}
        \caption{RT architecture.}
    \end{subfigure}
    \caption{Implemented DA and RT workflow architectures for the
    multi-agentic aggregator.}
    \label{fig:actor_flow}
\end{figure*}

As shown in Figure~\ref{fig:actor_flow}, the implementation is divided
into two event-driven workflows. The DA workflow creates the daily
reference plan, while the RT workflow repeatedly evaluates whether that
reference remains valid under updated market and fleet conditions.

Table~\ref{tab:agent_roles} summarizes the three supervisory agents
used in the implemented workflows. The implementation avoids
unconstrained free-text prompting by requiring domain-bounded
instructions and structured outputs that can be validated before they
affect the optimization engine.

\begin{table*}[t]
\centering
\caption{Agent roles in the implemented workflows.}
\label{tab:agent_roles}
\begin{adjustbox}{max width=\textwidth}
\begin{tabular}{L{2.5cm}L{5.2cm}L{7.8cm}}
\toprule
Agent & Main prompt content & Expected structured output \\
\midrule
Trigger Agent
    & Current context, DA plan summary, disturbance indicators,
      price deviation, energy-consumption deviation,
      service-delay indicators, and feasibility risk flags
    & Trigger decision $\delta_{\tau} \in \{\texttt{skip},
      \texttt{optimize}\}$ with rationale and confidence score \\
\addlinespace
Pricing Agent
    & Coordination mode, market conditions, average prices,
      recent outcomes, best-known multipliers, and pricing
      structures for aggregator--PTO trading
    & Price-guidance vector $\mathbf{u}_{\tau}$ containing bounded
      buy/sell multipliers per tariff period \\
\addlinespace
Evaluator Agent
    & Optimization result, feasibility status, PTO cost, aggregator
      revenue, V2G activity, disturbance indicators, rerun count,
      and acceptance criteria
    & Accept or rerun decision $a_{\tau}$, rationale $\ell_{\tau}$,
      confidence $\gamma_{\tau}$, and adjusted guidance request \\
\bottomrule
\end{tabular}
\end{adjustbox}
\end{table*}

The agent outputs are restricted to structured actions such as
\texttt{optimize}, \texttt{skip}, or \texttt{rerun}, bounded
price-guidance vectors, confidence levels, and acceptance decisions.
These restrictions are essential in a cyber-physical setting because
they prevent the agentic aggregator from acting as an unconstrained
controller. The framework is therefore best understood as a
two-timescale orchestration layer that adds context awareness, adaptive
triggering, and economic guidance around a formal scheduling model. The
architecture was implemented as two complementary event-driven
workflows using \texttt{n8n} as the orchestration environment.
Persistent state variables encoding the last accepted decision,
historical outcomes, and coordination-mode configuration are maintained
across timesteps.

\subsubsection{Day-Ahead and Real-Time Workflows}
\label{sec:da_rt_workflows}

The DA workflow constructs the daily reference plan. The Pricing Agent
first proposes buy and sell multipliers for each tariff period according
to the selected coordination mode. These multipliers are converted into
PTO-facing buy and sell tariffs by the optimization interface described
in Section~\ref{sec:pto_summary}. The optimizer then produces a fleet
charging and V2G schedule, and the Evaluator Agent either accepts the
result or requests a bounded revision with updated guidance. This loop
repeats up to $R_{\max}$ times, and the best accepted plan is exported
as the DA reference schedule $\mathbf{s}_{\mathrm{DA}}^{*}$.
Algorithm~\ref{alg:da_workflow} details this sequence.

\begin{algorithm}[ht]
\caption{DA planning workflow}
\label{alg:da_workflow}
\begin{algorithmic}[1]
    \State Fetch fleet, charger, trip, disturbance, and market data for day $d$
    \State Select coordination mode
           $m \in \{\text{operational-based},\text{profit-based}\}$
    \State Build decision context $\mathbf{c}_{d}^{\mathrm{DA}} =
           \Psi(\mathbf{f}_{d}, \mathbf{g}_{d}, \mathbf{h}_{d}, m)$,
           loading historical memory $\mathbf{h}_{d}$ from prior
           accepted outcomes
    \For{$r = 0,\ldots,R_{\max}$}
        \State Pricing Agent proposes
               $\mathbf{u}_{d,r}^{\mathrm{DA}}=
               \{\alpha_{p,r}^{+},\alpha_{p,r}^{-}\}_{p\in\mathcal{P}}$
        \State Convert $\mathbf{u}_{d,r}^{\mathrm{DA}}$ into tariffs
               $\{\rho_{p,r}^{+},\rho_{p,r}^{-}\}_{p\in\mathcal{P}}$
        \State Solve PTO optimization and return schedule $\mathbf{s}_{d,r}^{*}$
        \State Compute KPIs: feasibility, PTO cost, aggregator revenue,
               V2G energy, and terminal SOC
        \State Evaluator Agent returns $(a_{d,r},\ell_{d,r},\gamma_{d,r})$
        \If{$a_{d,r}=\texttt{accept}$ \textbf{or} $r=R_{\max}$}
            \State Save $\mathbf{s}_{d,r}^{*}$, KPIs, tariffs, and agent rationale
            \State Export accepted DA plan as $\mathbf{s}_{\mathrm{DA}}^{*}$
            \State \textbf{break}
        \Else
            \State Update pricing memory and prepare constrained rerun guidance
        \EndIf
    \EndFor
\end{algorithmic}
\end{algorithm}

The RT workflow updates this plan only when operating conditions
justify intervention. At each 30-minute interval, the Trigger Agent
compares the current fleet and market state against the active
reference schedule. If no material deviation is detected, the previous
schedule is retained. If re-optimization is triggered, the Pricing Agent
and Evaluator Agent are invoked under the same logic as in the DA
workflow, but the optimizer runs only over the remaining horizon from
the current timestep. The updated optimization uses observed SOC,
service state, electricity prices, and revised route-energy
requirements. Algorithm~\ref{alg:rt_workflow} details this sequence.

\begin{algorithm}[ht]
\caption{RT adaptation workflow}
\label{alg:rt_workflow}
\begin{algorithmic}[1]
    \State Load last accepted plan $\mathbf{s}^{*}_{\tau-1}$ and DA reference
           $\mathbf{s}^{*}_{\mathrm{DA}}$
    \State Fetch current fleet, charger, disturbance, and market data
    \State Build current context $\mathbf{c}_{\tau}^{\mathrm{RT}} =
           \Psi(\mathbf{f}_{\tau}, \mathbf{g}_{\tau},
           \mathbf{h}_{\tau}, m)$
    \State Trigger Agent evaluates $\mathcal{T}_{\tau}$ and returns $\delta_{\tau}$
    \If{$\delta_{\tau}=\texttt{skip}$}
        \State Set $\mathbf{s}^{*}_{\tau} \leftarrow \mathbf{s}^{*}_{\tau-1}$
               and log trigger rationale
    \Else
        \For{$r = 0,\ldots,R_{\max}$}
            \State Pricing Agent proposes updated guidance
                   $\mathbf{u}_{\tau,r}^{\mathrm{RT}}$
            \State Solve PTO optimization from the current state and return
                   $\mathbf{s}_{\tau,r}^{*}$
            \State Compute RT KPIs and deviation from $\mathbf{s}^{*}_{\mathrm{DA}}$
            \State Evaluator Agent returns $(a_{\tau,r},\ell_{\tau,r},\gamma_{\tau,r})$
            \If{$a_{\tau,r}=\texttt{accept}$ \textbf{or} $r=R_{\max}$}
                \State Save $\mathbf{s}^{*}_{\tau} \leftarrow \mathbf{s}_{\tau,r}^{*}$
                       and update memory
                \State \textbf{break}
            \Else
                \State Restrict the next guidance update using evaluator feedback
            \EndIf
        \EndFor
    \EndIf
    \State Use $\mathbf{s}^{*}_{\tau}$ as the reference at timestep $\tau+1$
\end{algorithmic}
\end{algorithm}

The formal mathematical definitions of the agent policies and their
connection to the PTO scheduling problem are given in
Appendix~\ref{sec:agentic_logic}.

\subsubsection{Aggregator Coordination Modes: Profit-based and
Operational-based}
\label{sec:behavior_modes}

The agentic aggregator can be configured according to different
coordination modes that define its economic posture when it proposes
price guidance and evaluates candidate schedules. Two modes are
considered: profit-based and operational-based.

\begin{itemize}
    \item In the \emph{profit-based} mode, the aggregator behaves as a
    commercially oriented intermediary. The Pricing Agent is allowed to
    propose higher charging-price multipliers when the PTO buys energy
    and lower V2G compensation multipliers when the PTO exports energy.
    This mode represents an upper-bound case for per-fleet value
    capture, since it reveals how much revenue the aggregator can obtain
    when market margin is the dominant coordination objective. It also
    exposes the associated PTO-side cost burden, which would need to be
    justified through contracts, reliability guarantees, or explicit
    value-sharing rules.

    \item In the \emph{operational-based} mode, the aggregator behaves
    as a fleet-oriented coordinator. The objective is not to eliminate
    aggregator revenue, but to make grid participation compatible with
    public transport operation. The Pricing Agent therefore favors lower
    charging-price multipliers and higher V2G compensation multipliers,
    reducing the PTO's exposure to electricity-price markups and
    returning more value to the fleet when it provides flexibility. This
    mode reflects an aggregator operated by a public institution, a
    regulated platform, or a private provider whose business model
    depends on long-term fleet participation and portfolio scale.
\end{itemize}

For each tariff period $p$, the buy multiplier $\alpha_p^{+}$ and the
sell multiplier $\alpha_p^{-}$ proposed by the Pricing Agent are
bounded by mode-dependent ranges:
\begin{equation}
    \underline{\alpha}_{m}^{+} \le \alpha_{p}^{+} \le \overline{\alpha}_{m}^{+},
    \qquad
    \underline{\alpha}_{m}^{-} \le \alpha_{p}^{-} \le \overline{\alpha}_{m}^{-},
    \qquad \forall p\in\mathcal{P},
    \label{eq:mode_bounds}
\end{equation}
where $m$ denotes the selected coordination mode. In the
operational-based mode, these bounds restrict the aggregator to
PTO-compatible pricing ranges. In the profit-based mode, they allow
wider margins for aggregator value capture. The bounds therefore make
the aggregator's economic posture explicit while keeping the
agent-generated tariffs within predefined and auditable limits. The
conversion from these multipliers to PTO-facing tariffs is given in
Section~\ref{sec:pto_summary}.

After a candidate schedule is produced by the optimizer, the Evaluator
Agent assesses whether the result is consistent with the selected mode
through a mode-dependent supervisory score:
\begin{equation}
    J_m(\mathbf{u}_{\tau},\mathbf{s}_{\tau}^{*})
    =
    \lambda_{m}^{A}\Pi_{\tau}^{\mathrm{Agg}}
    - \lambda_{m}^{P}C_{\tau}^{\mathrm{PTO}}
    - \lambda_{m}^{R}R_{\tau},
    \label{eq:behavior_score}
\end{equation}
where $\Pi_{\tau}^{\mathrm{Agg}}$ is aggregator revenue,
$C_{\tau}^{\mathrm{PTO}}$ is PTO operating cost, and $R_{\tau}$ is a
supervisory operational-risk score.\footnote{The term $R_{\tau}$ is not
an additional constraint in the PTO optimization model. It is an
evaluation metric used by the agentic layer to discourage schedules
with weak service robustness, low terminal SOC margins, or unresolved
operational warnings.}
The revenue and cost terms are defined by the optimization interface in
Section~\ref{sec:pto_summary} and by the formal supervisory flow in
Appendix~\ref{sec:agentic_logic}. The decomposition of $R_{\tau}$ is
given by
\begin{equation}
    R_{\tau}
    =
    \omega_{\mathrm{SOC}}R_{\tau}^{\mathrm{SOC}}
    +
    \omega_{\mathrm{end}}R_{\tau}^{\mathrm{end}}
    +
    \omega_{\mathrm{serv}}R_{\tau}^{\mathrm{serv}},
    \label{eq:risk_components}
\end{equation}
where $R_{\tau}^{\mathrm{SOC}}$ captures low-SOC exposure during the
remaining horizon, $R_{\tau}^{\mathrm{end}}$ captures end-of-day
readiness risk, and $R_{\tau}^{\mathrm{serv}}$ captures
service-feasibility warnings. In the operational-based mode,
$\lambda_m^P$ and $\lambda_m^R$ are larger relative to $\lambda_m^A$,
so the supervisory layer favors lower PTO cost and more robust
schedules. In the profit-based mode, $\lambda_m^A$ is larger, so higher
aggregator revenue receives greater weight, provided that feasibility
remains acceptable.

\subsubsection{Event-Triggered Re-Optimization Logic}
\label{sec:event_trigger}

The event-triggered logic determines when the RT workflow should rerun
the PTO optimization model. At each RT interval $\tau$, the Trigger
Agent compares the observed operating context with the assumptions
embedded in the active reference plan. Three disturbance classes are
monitored. The first class includes electricity-price deviations, which
may change the economic value of charging or V2G export. The second
class includes route-energy deviations, which affect SOC trajectories
and may reduce the energy margin available for service or flexibility.
The third class includes service-timing disturbances, such as early or
late arrivals, which modify depot dwell times and charging
opportunities.

These signals are converted into normalized indicators and summarized
by a compact trigger score:
\begin{equation}
    \mathcal{T}_{\tau}
    =
    \omega_{p}\Delta^{p}_{\tau}
    +
    \omega_{e}\Delta^{e}_{\tau}
    +
    \omega_{d}\Delta^{d}_{\tau},
    \label{eq:trigger_score}
\end{equation}
where $\Delta^{p}_{\tau}$ measures the normalized price deviation,
$\Delta^{e}_{\tau}$ measures the normalized route-energy deviation, and
$\Delta^{d}_{\tau}$ measures service-delay severity. The weights
$\omega_p$, $\omega_e$, and $\omega_d$ reflect the relative importance
of market, energy, and service-timing deviations in the triggering
logic. A new RT optimization run is activated when the aggregate trigger
score exceeds a predefined threshold or when a hard feasibility warning
is detected:
\begin{equation}
    \mathcal{T}_{\tau} \ge \Theta
    \quad \text{or} \quad
    \mathrm{FR}_{\tau}=1.
    \label{eq:trigger_rule}
\end{equation}
Here, $\Theta$ is the trigger threshold and $\mathrm{FR}_{\tau}$ is a
binary flag identifying cases in which the current reference plan should
not continue without intervention. This flag is defined as
\begin{equation}
\begin{split}
\mathrm{FR}_{\tau}
&=
\mathbb{I}\big[
\exists k\in\mathcal{K}:
e_{k,\tau}^{\mathrm{obs}} < E_k^{\min}
\ \lor\
e_{k,T}^{\mathrm{pred}} < E_k^{\mathrm{end}}
\\
&\hspace{3.8cm}
\lor\
\mathrm{TR}_{k,\tau}=1
\big],
\end{split}
\label{eq:feasibility_risk}
\end{equation}
where $\mathbb{I}[\cdot]$ is an indicator function,
$e_{k,\tau}^{\mathrm{obs}}$ is the observed stored energy of bus $k$ at
interval $\tau$, $e_{k,T}^{\mathrm{pred}}$ is the terminal stored energy
expected if the current reference schedule is maintained, and
$\mathrm{TR}_{k,\tau}$ indicates whether bus $k$ is at risk of missing
an active or upcoming service assignment.

If neither condition in Eq.~\eqref{eq:trigger_rule} is satisfied, the
RT workflow keeps the current reference schedule and no optimization is
solved. If the trigger condition is satisfied, the PTO optimization
model is rerun from the current interval to the end of the operating day
using the observed SOC, service state, updated electricity prices, and
observed route-level energy requirements. The accepted RT solution then
becomes the new reference schedule for subsequent intervals.
Figure~\ref{fig:rt_logic} exemplifies this dynamic.


\subsection{PTO Scheduling and Optimization Interface}
\label{sec:pto_summary}

The agents described above govern a public transport operator (PTO)
charging-scheduling model. This subsection states the compact
optimization interface used in the body of the paper. The complete
constraint set, the RT re-optimization variant, and the formal
connection between the supervisory layer and the optimizer are provided
in ~\ref{sec:bilevel}.

Let $p \in \mathcal{P}$ index tariff periods, $t \in \mathcal{T}$ time
intervals, $k \in \mathcal{K}$ buses, $n \in \mathcal{N}$ chargers, and
$i \in \mathcal{I}$ routes. The PTO decides route assignments
$b_{k,i,t}$, charging $x_{k,n,t}$, V2G discharging $y_{k,n,t}$, stored
energy $e_{k,t}$, and the aggregate depot power exchanged with the
grid, $w_t^{+}$ for import and $w_t^{-}$ for export.

The Pricing Agent proposes period multipliers $\alpha_p^{+}$ and
$\alpha_p^{-}$, which are mapped into the PTO-facing buy and sell
tariffs by scaling the average grid price $\bar{\pi}_p^{g}$:
\begin{equation}
\rho_p^{+} = \alpha_p^{+}\,\bar{\pi}_p^{g},
\qquad
\rho_p^{-} = \alpha_p^{-}\,\bar{\pi}_p^{g},
\qquad \forall p \in \mathcal{P}.
\label{eq:tariff_map}
\end{equation}
Given these tariffs, the PTO selects a schedule $\mathbf{s}^{*}$ that
minimizes its net energy-trading cost:
\begin{equation}
\mathbf{s}^{*}
\in
\arg\min_{\mathbf{s}\in\mathcal{S}}
C^{\mathrm{PTO}}(\mathbf{s}),
\label{eq:pto_argmin}
\end{equation}
where
\begin{equation}
C^{\mathrm{PTO}}(\mathbf{s})
=
\sum_{t \in \mathcal{T}}
\left[
\rho_{p(t)}^{+} w_t^{+}
-
\rho_{p(t)}^{-} w_t^{-}
\right]\Delta t,
\label{eq:pto_obj}
\end{equation}
and $\mathcal{S}$ is the feasible set defined by the operational
constraints.

The binding energy dynamics are given by the per-bus SOC balance:
\begin{align}
\begin{split}
e_{k,t+1}
&=
e_{k,t}
+
\sum_{n \in \mathcal{N}}
\eta_n^{\mathrm{ch}} P_n^{\mathrm{ch}} x_{k,n,t}\Delta t
\\
&\quad -
\sum_{n \in \mathcal{N}}
\tfrac{1}{\eta_n^{\mathrm{dis}}}
P_n^{\mathrm{dis}} y_{k,n,t}\Delta t
-
\sum_{i \in \mathcal{I}}\xi_i b_{k,i,t},
\end{split}
\label{eq:soc_dyn}
\end{align}
subject to battery limits and terminal-energy requirements,
\begin{equation}
E_k^{\min}\le e_{k,t}\le E_k^{\max}, \qquad
e_{k,0}=E_k^{0}, \qquad
e_{k,T}\ge E_k^{\mathrm{end}}.
\label{eq:compact_battery_limits}
\end{equation}
The feasible set also includes route--charger exclusivity, so that a
bus cannot charge and serve a route in the same interval; per-charger
occupancy constraints; depot-power aggregation constraints,
\begin{equation}
w_t^{+}
=
\sum_{k\in\mathcal{K}}\sum_{n\in\mathcal{N}}
P_n^{\mathrm{ch}}x_{k,n,t},
\qquad
w_t^{-}
=
\sum_{k\in\mathcal{K}}\sum_{n\in\mathcal{N}}
P_n^{\mathrm{dis}}y_{k,n,t};
\label{eq:compact_depot_power}
\end{equation}
and depot exchange limits,
\begin{equation}
0\le w_t^{+}\le \overline{W},
\qquad
0\le w_t^{-}\le \overline{W}.
\label{eq:compact_depot_limits}
\end{equation}
The full mathematical formulation is deferred to
~\ref{sec:bilevel} to keep the body of the methodology focused
on the agentic coordination logic.

\section{Case Study Description}
\label{sec:case}

The case study evaluates the proposed agentic aggregator in a controlled electric-bus depot setting. The system includes the main elements required for fleet-grid coordination: buses, battery limits, depot chargers, route-service windows, route-level energy requirements, and time-varying electricity prices. The analysis is organized in two stages. First, the DA evaluation defines the reference schedule under nominal conditions. Second, the RT evaluation introduces controlled disturbances in service timing, route energy consumption, and electricity prices. In the following, we describe the bus-system characteristics and the DA and RT evaluation scenarios, respectively.

\subsection{Fleet and Charging Configuration}

The fleet and infrastructure configuration is summarized in Table~\ref{tab:case_fleet}. The case study involves a depot-based electric bus system with $8$ buses, each with a battery capacity of $365$~kWh and an initial SOC of $20\%$, served by $8$ chargers rated at $200$~kW. At the depot level, the charging process is coordinated over $48$ half-hour intervals, and the case-study instance is evaluated under DA conditions before moving to RT adjustments. This setting remains representative of a realistic fleet depot in which charger ratings, aggregate energy availability, and service schedules jointly constrain operations.

\begin{figure*}[t]
    \centering
    \includegraphics[width=0.65\linewidth]{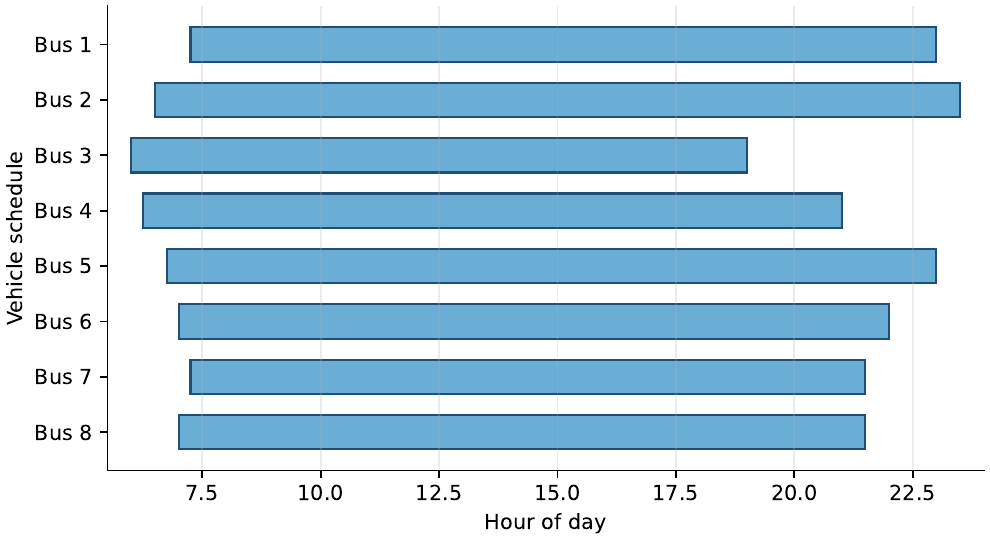}
    \caption{Daily service windows extracted from the trip-time input sheet.}
    \label{fig:case_trip_windows}
\end{figure*}

The operational side is defined through $8$ route blocks with heterogeneous service windows. According to the trip-time sheet, route activity starts between 05:30 and 07:00 and finishes between 18:30 and 23:00, leaving different charging opportunities before, between, and after service blocks. Figure~\ref{fig:case_trip_windows} shows these windows for each vehicle. The corresponding nominal route energy-consumption values range from $0.859$ to $0.923$~kWh/km, with an average of approximately $0.891$~kWh/km, which is sufficiently demanding to make both dwell-time compression and route-energy deviations operationally consequential.

\begin{figure*}[t]
	\centering
	\includegraphics[width=0.8\linewidth]{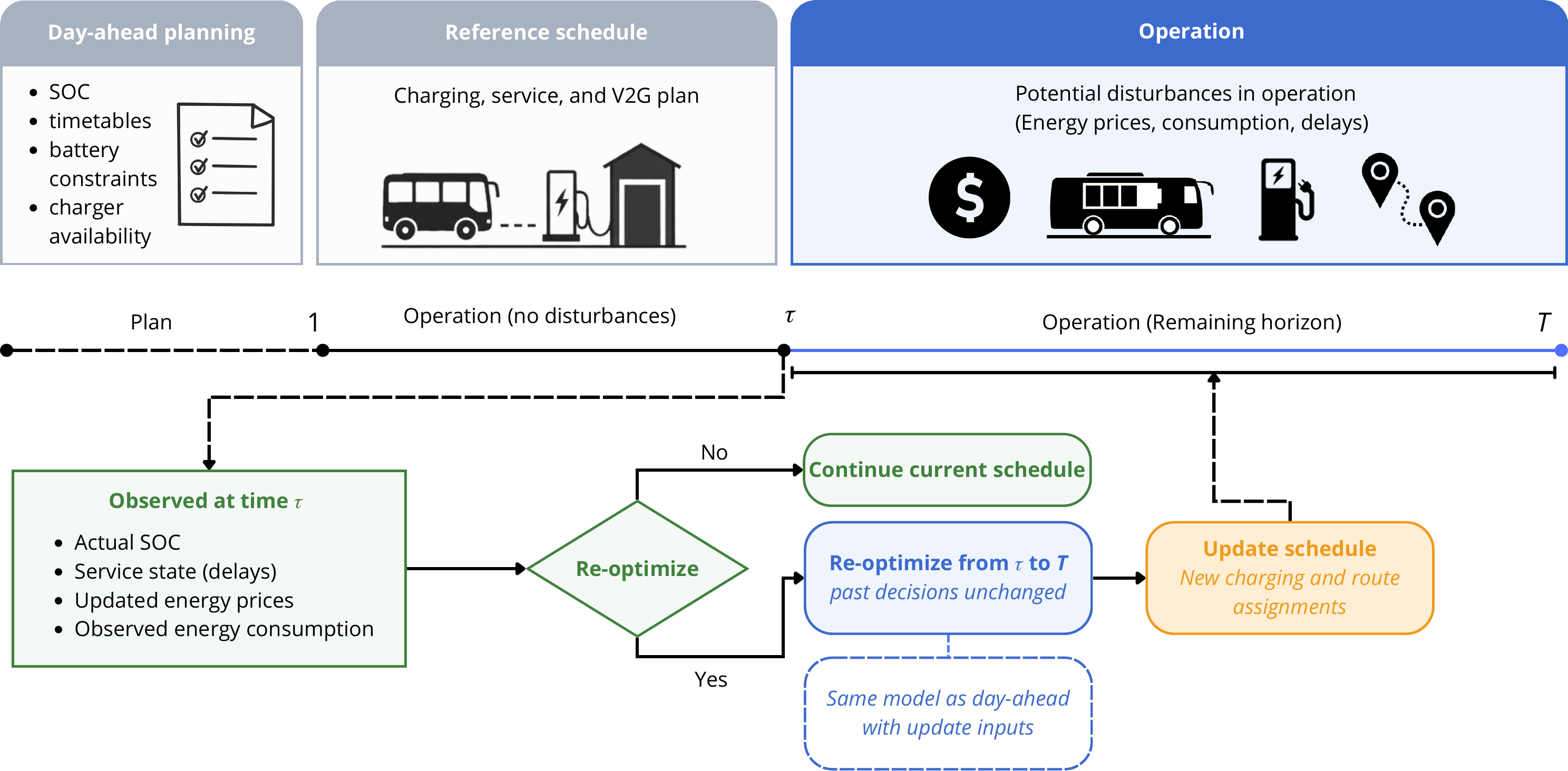}
	\caption{Real-time optimization logic.}
	\label{fig:rt_logic}
\end{figure*}

 \begin{table}[t]
\centering
\caption{Main fleet, infrastructure, and horizon parameters in the case study.}
\label{tab:case_fleet}
\begin{tabular}{l c}
\hline
Parameter & Value \\
\hline
Number of buses & 8 \\
Battery capacity per bus & 365 kWh \\
Initial SOC per bus & 20\% \\
Number of chargers & 8 \\
Charger rating & 200 kW \\
Routes & 8 \\
Scheduling horizon & 24 h \\
Time discretization & 30 min (48 intervals) \\
\hline
\end{tabular}
\end{table}

The electricity-market inputs are summarized in Table~\ref{tab:case_prices}. Spot-market prices range from approximately 0.067 to 0.122 EUR/kWh, with a daily mean of about 0.090 EUR/kWh. In the DA analysis, this profile is the common benchmark price seen by the non-agentic scenarios and the reference against which the multi-agentic aggregator defines PTO buy and sell tariffs. In the RT analysis, the same base profile serves as the anchor for determining whether the system remained on the DA tariff or switched to a revised RT tariff after an accepted re-optimization.

\begin{table}[t]
\centering
\caption{Electricity-price characteristics of the case-study instance.}
\label{tab:case_prices}
\begin{tabular}{l c}
\hline
Price parameter & Value \\
\hline
Spot price range & 0.067--0.122 EUR/kWh \\
Average spot price & 0.090 EUR/kWh \\
Number of intervals & 48 \\
Interval duration & 30 min \\
\hline
\end{tabular}
\end{table}

\subsection{Day-Ahead Planning Strategies}
\label{sec:da_scenarios}

The DA analysis is organized around four strategies that progressively move from non-agentic charging benchmarks toward agent-guided aggregator coordination. These strategies are summarized in Table~\ref{tab:da_scenarios}.

The first two strategies provide operational baselines: a rule-based "dumb charging" policy and an optimization-based smart-charging policy without V2G participation. The last two strategies activate the pricing role of the multi-agentic aggregator under profit-based and operational-based coordination. They aim to show how different aggregator postures affect the balance between grid-facing flexibility value and PTO-facing operating needs.

\subsection{Real-Time Evaluation Scenarios}
\label{sec:rt_scenarios}

The RT evaluation examines how the multi-agentic aggregator responds when observed operating conditions diverge from the accepted DA reference plan, $\mathbf{s}_{\mathrm{DA}}^{*}$. Controlled perturbations are introduced in service timing, route energy consumption, electricity prices, and selected combined cases, representing typical mismatches between planning assumptions and realized electric-bus operation. Table~\ref{tab:rt_scenarios} summarizes the evaluated RT disturbance scenarios.

The experiment uses structured input files to emulate 30-minute operational updates. At each RT interval, the workflow builds the observed system context and applies the trigger logic introduced in Section~\ref{sec:event_trigger}. If $\delta_{\tau}=\texttt{skip}$, the active schedule is maintained and the monitoring record is updated. If $\delta_{\tau}=\texttt{optimize}$, the optimization model is solved from the current observed state, and the resulting schedule is evaluated, accepted, or sent to a bounded rerun. 

\section{Results and Discussion}
\label{sec:results}

\subsection{Day-Ahead Optimization Results}
\label{sec:da_results}

Table~\ref{tab:da_all_results} compares the four DA scenarios and shows a clear progression from conservative charging to flexibility-based operation.
\begin{table*}[t]
	\centering
	\caption{DA strategies considered in the evaluation design.}
	\label{tab:da_scenarios}
	\small
	\setlength{\tabcolsep}{4pt}
	\renewcommand{\arraystretch}{1.08}
	\begin{tabularx}{\textwidth}{@{}p{0.06\textwidth} p{0.22\textwidth} p{0.07\textwidth} X@{}}
		\toprule
		ID & Strategy & V2G & Main purpose \\
		\midrule
		S1 & Dumb charging & Off & Rule-based benchmark that charges buses as early as possible without market-responsive optimization. \\
		S2 & Smart charging (no V2G) & Off & Deterministic cost-minimizing charging benchmark using spot-market prices directly. \\
		S3 & Profit-based aggregator & On & Agent-guided pricing case in which the aggregator prioritizes per-fleet revenue and stronger tariff margins while preserving feasibility. \\
		S4 & Operational-based aggregator & On & Agent-guided pricing case in which the aggregator prioritizes PTO-compatible flexibility provision and lower tariff exposure. \\
		\bottomrule
	\end{tabularx}
\end{table*}

\begin{table*}[t]
	\centering
	\caption{RT scenarios in the evaluation design. Each scenario is analyzed using profit-based (S3) and operational-based (S4) strategies.}
	\label{tab:rt_scenarios}
	\begin{tabularx}{\linewidth}{@{}L{1.6cm}L{2.0cm}L{2.4cm}L{2.0cm}X@{}}
		\toprule
		Scenario & Family & Disturbance & Time & Description \\
		\midrule
		
		D+30 beg.
		& \multirow[t]{4}{1.0cm}{Service timing}
		& $+30$ min delay
		& 04:30--09:00
		& Early-day delay compresses depot charging windows and reduces V2G export opportunities. \\
		D-30 beg.
		&
		& $-30$ min early return
		& 04:30--09:00
		& Early-day early return creates unplanned dwell time; tests whether the schedule recovers flexibility. \\
		D+30 end
		&
		& $+30$ min delay
		& 17:30--24:00
		& Late-day delay shortens the end-of-day charging window and threatens terminal SOC reserve. \\
		D-30 end
		&
		& $-30$ min early return
		& 17:30--24:00
		& Late-day early return frees time near the end of the horizon; tests V2G export response. \\
		
		\midrule
		E+50
		& \multirow[t]{2}{1.0cm}{Route energy}
		& $+50\%$ kWh/km, all buses
		& 06:00--20:00
		& Higher traction demand raises grid purchases, depletes SOC reserve, and suppresses V2G. \\
		E-50
		&
		& $-50\%$ kWh/km, all buses
		& 06:00--20:00
		& Lower traction demand frees surplus SOC; tests whether the aggregator expands V2G export. \\
		
		\midrule
		P+25
		& \multirow[t]{4}{1.0cm}{Electricity price}
		& $+25\%$ spot price
		& 02:30--05:00
		& Moderate positive price shock; tests tariff-guided reduction of charging during the window. \\
		P-25
		&
		& $-25\%$ spot price
		& 02:30--05:00
		& Moderate negative price shock; tests whether the aggregator shifts charging into the low-price window. \\
		P+50
		&
		& $+50\%$ spot price
		& 02:30--05:00
		& Strong positive price shock; tests the upper bound of tariff exposure and V2G suppression. \\
		P-50
		&
		& $-50\%$ spot price
		& 02:30--05:00
		& Strong negative price shock; tests maximum charging shift and V2G compensation response. \\
		
		\midrule
		C-Seq
		& \multirow[t]{4}{1.0cm}{Combined}
		& $P+50$ then $E+50$ then $D+30$
		& 02:30--05:00, 12:30--15:00, 17:30--24:00
		& Sequential arrival of market, energy, and timing stressors; each disturbance resolved before the next arrives. \\
		C-All 5--48
		&
		& $P+50$, $E+50$, $D+30$ simultaneous
		& 02:30--24:00
		& Full-day simultaneous stress; all three disturbance types active across most of the operating horizon. \\
		C-All 5--25
		&
		& $P+50$, $E+50$, $D+30$ simultaneous
		& 02:30--12:30
		& Early/midday simultaneous stress; disturbances resolve shortly after midday, leaving a clean late horizon. \\
		C-All 20--48
		&
		& $P+50$, $E+50$, $D+30$ simultaneous
		& 10:00--24:00
		& Late-morning-to-end-of-day simultaneous stress; disturbances begin after the morning period with limited remaining horizon to recover. \\
		\bottomrule
	\end{tabularx}
\end{table*}

\begin{table*}[t]
\centering
\caption{DA results across the four evaluated scenarios using SOC-derived energy accounting.}
\label{tab:da_all_results}
\begin{adjustbox}{max width=\textwidth}
\begin{tabular}{lcccccccc}
\toprule
\makecell[l]{Scenario} & \makecell{PTO cost\\(EUR/day)} & \makecell{Aggregator\\revenue\\(EUR/day)} & \makecell{Bought\\(kWh/day)} & \makecell{Sold\\(kWh/day)} & \makecell{Avg. buy\\price\\(EUR/kWh)} & \makecell{Avg. sell\\price\\(EUR/kWh)} & \makecell{Min.\\SOC\\(\%)} & \makecell{End avg.\\SOC\\(\%)} \\
\midrule
Dumb charging (S1) & 218.10 & 0.00 & 2400.0 & 0.0 & 0.0909 & -- & 20.00 & 52.38 \\
Smart charging, no V2G (S2) & 130.47 & 0.00 & 1600.0 & 0.0 & 0.0815 & -- & 20.00 & 25.61 \\
Profit-based aggregator (S3) & 140.59 & 20.30 & 1900.0 & 300.0 & 0.0874 & 0.0847 & 20.00 & 23.44 \\
Operational-based aggregator (S4) & 118.91 & 2.39 & 2000.0 & 400.0 & 0.0826 & 0.1158 & 20.00 & 22.72 \\
\bottomrule
\end{tabular}
\end{adjustbox}
\end{table*}

Dumb charging (S1) provides the conservative baseline, with the highest PTO cost (218.10 EUR/day), the largest grid purchase (2400 kWh/day), and the highest average terminal SOC (52.38\%). Smart charging without V2G (S2) reduces the PTO cost by 40.2\%, to 130.47~EUR/day, by shifting charging to lower-cost periods and reducing purchases to 1600~kWh/day, although the terminal SOC falls to 25.61\%. The agentic cases (S3 and S4) add bidirectional grid exchange, but they express different reconciliation policies through both tariffs and dispatch. The profit-based mode (S3) buys 1900~kWh/day and sells 300~kWh/day, raising aggregator revenue to 20.30~EUR/day but increasing PTO cost to 140.59~EUR/day. The operational-based mode (S4) buys 2000~kWh/day and sells 400~kWh/day, reaching the lowest PTO cost, 118.91~EUR/day, while reducing aggregator revenue to 2.39~EUR/day.

\begin{figure*}[t]
    \centering
    \begin{subfigure}[t]{0.55\textwidth}
        \centering
        \includegraphics[width=\textwidth]{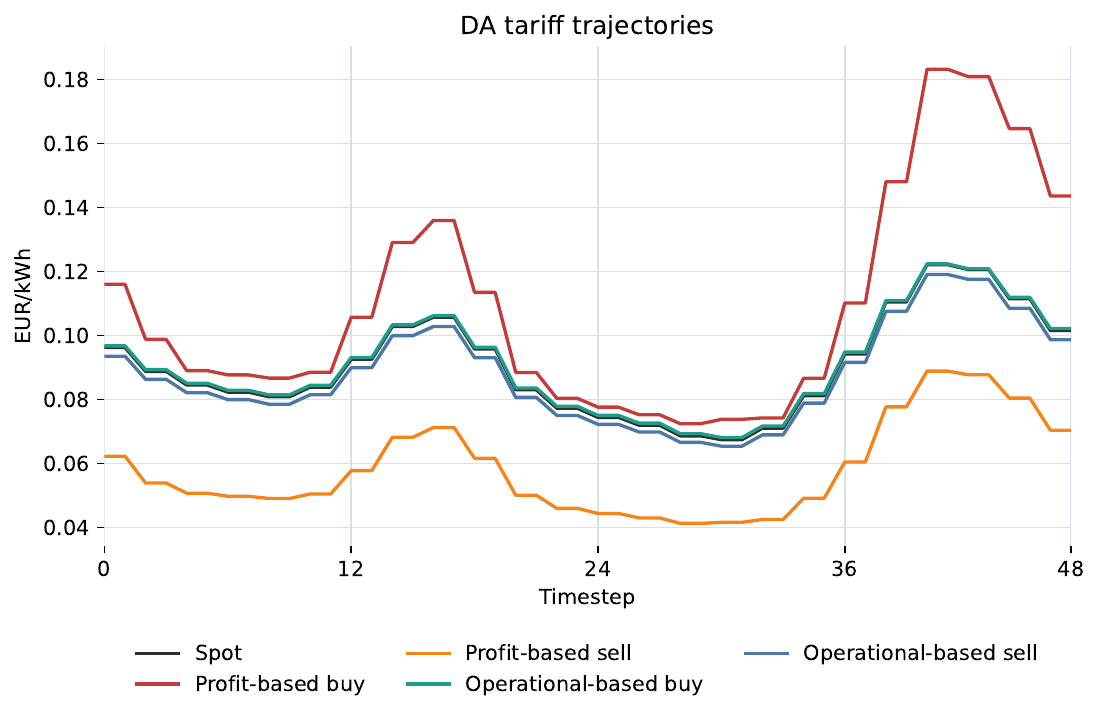}
        \caption{Time-varying buy and sell tariffs.}
        \label{fig:da_agentic_prices_timeseries}
    \end{subfigure}
    \hfill
    \begin{subfigure}[t]{0.4\textwidth}
        \centering
        \includegraphics[width=\textwidth]{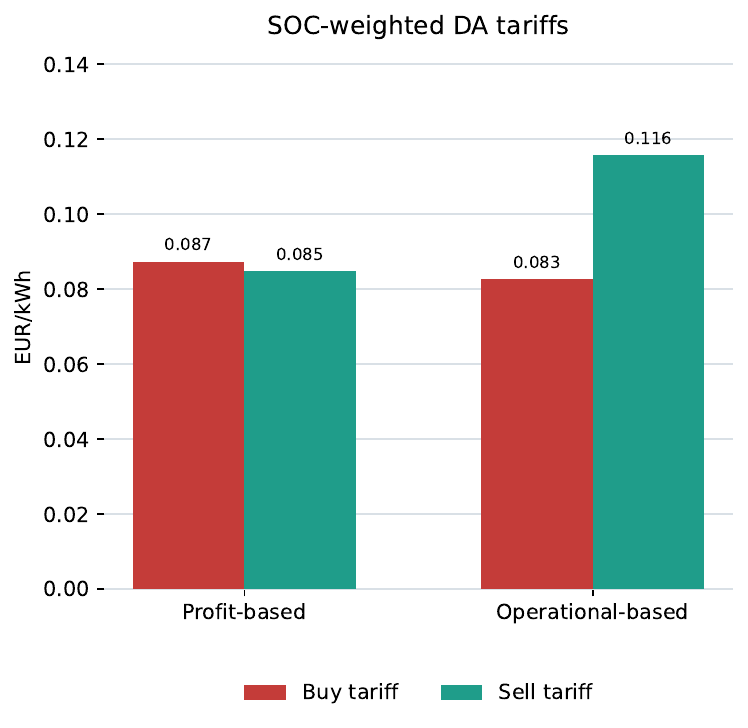}
        \caption{SOC-weighted average tariffs.}
        \label{fig:da_agentic_prices_average}
    \end{subfigure}
    \caption{DA tariff comparison for the agentic scenarios. Panel~(a) shows the time-varying buy and sell price curves. Panel~(b) reports the corresponding SOC-weighted average buy and sell price levels.}
    \label{fig:da_agentic_prices}
\end{figure*}

Figure~\ref{fig:da_agentic_prices} explains how the two agentic cases create different financial outcomes through the tariff vectors selected by the Pricing Agent.

\begin{figure*}[t]
    \centering
    \begin{subfigure}[t]{0.49\textwidth}
        \centering
        \includegraphics[width=\textwidth]{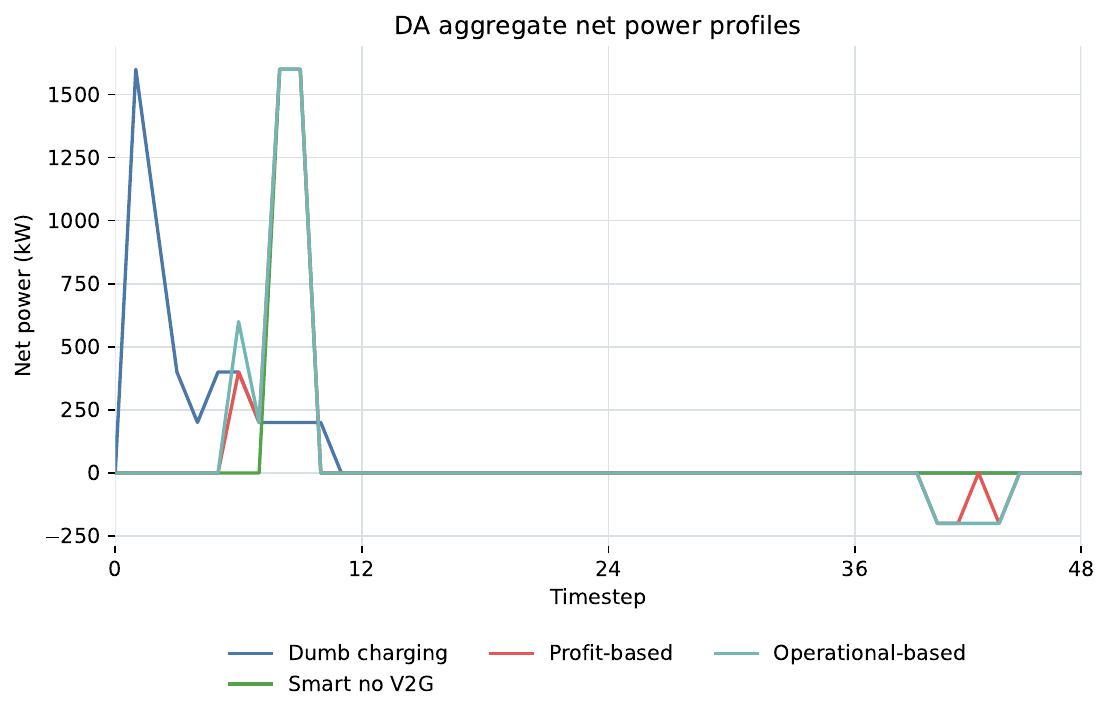}
        \caption{Aggregated charging and discharging power profiles.}
        \label{fig:da_power_profiles}
    \end{subfigure}
    \hfill
    \begin{subfigure}[t]{0.49\textwidth}
        \centering
        \includegraphics[width=\textwidth]{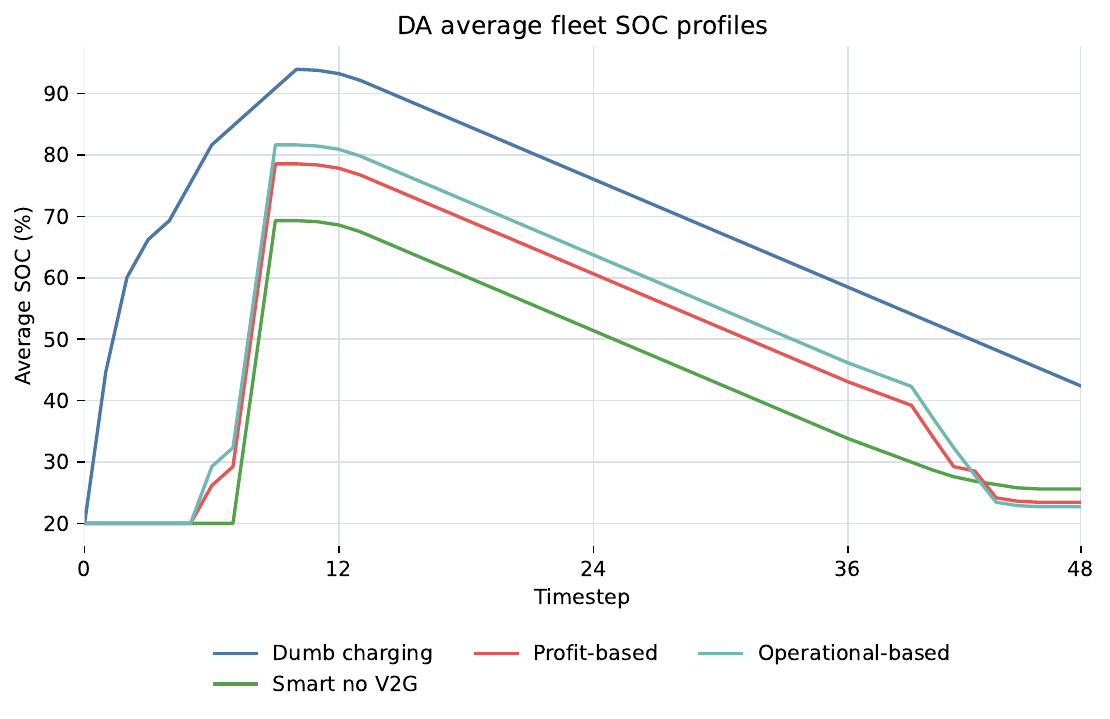}
        \caption{Average fleet SOC trajectories.}
        \label{fig:da_energy_profiles}
    \end{subfigure}
    \caption{DA operating profiles. Panel~(a) compares aggregate power profiles. Panel~(b) compares the corresponding average fleet SOC trajectories.}
    \label{fig:da_physical_profiles}
\end{figure*}

The time-varying tariff trajectories show that the profit-based mode maintains a pricing structure that is more favorable to the aggregator, while the operational-based mode keeps the charging tariff close to the spot price and increases the V2G compensation. The weighted average prices make this contrast clearer: the operational-based case combines a lower average buy tariff (0.0826 EUR/kWh) with a substantially higher sell-back tariff (0.1158 EUR/kWh), whereas the profit-based case combines a higher buy tariff (0.0874 EUR/kWh) with a lower sell-back tariff (0.0847 EUR/kWh). The agentic layer therefore modifies how the economic value of flexibility is distributed between the PTO and the aggregator. The power and energy profiles in Figure~\ref{fig:da_physical_profiles} show the operational behavior behind these results.

In S1, charging is concentrated early in the horizon and remains purely unidirectional. This explains the large energy purchase and the high terminal SOC observed. In S2, charging is concentrated in a narrower time window, indicating that the optimizer selects more favorable charging periods instead of charging immediately whenever buses are available. The agentic cases, S3 and S4, introduce discharge events near the end of the horizon, but the operational-based case exports more energy than the profit-based case. The SOC trajectories provide an additional operational interpretation. All scenarios respect the 20\% lower SOC bound, but they use the available battery reserve differently. S1 keeps the largest energy buffer, ending at 52.38\% average SOC. S2 reduces this buffer to 25.61\%, which reflects a more efficient use of the battery capacity for cost minimization. S3 ends at 23.44\% average SOC after 300~kWh/day of V2G export, whereas S4 ends at 22.72\% after 400~kWh/day of export. The operational-based DA plan therefore gives the PTO both lower tariff exposure and a larger sell-back opportunity, while the profit-based plan retains more margin for the aggregator from a smaller export volume.

These results indicate that the main economic gain comes first from optimized charging, which reduces unnecessary energy purchases and shifts charging toward lower-cost periods. V2G adds a second layer of value by allowing the fleet to export energy, but the benefit depends strongly on the coordination mode selected by the aggregator. The profit-based mode favors aggregator revenue through a larger tariff spread, while the operational-based mode favors PTO-compatible participation through lower buy prices, higher sell-back compensation, and greater export. This point is important for the RT analysis, because the accepted DA plan establishes the expected cost, the aggregator revenue target, and the SOC margin available to absorb RT deviations.

\subsection{Real-Time Disturbance Results}
\label{sec:rt_results}

For consistency, RT deltas are computed against the DA reference of the
corresponding coordination mode, and the applied tariffs are reconstructed at
the timestep level: when no accepted RT tariff update is active, the DA
multiplier vector is used; once the RT workflow accepts an updated vector, the
executed tariff applies from that timestep onward. For price-disturbance cases
the market price is additionally scaled during the disturbance window. The
weighted buy and sell tariffs reported below follow
Eq.~\eqref{eq:rt_weighted_tariffs}:
\begin{equation}
\bar{\rho}^{+}
=
\frac{\sum_t w_t^{+}\rho_{p(t)}^{+}\Delta t}
{\sum_t w_t^{+}\Delta t},
\qquad
\bar{\rho}^{-}
=
\frac{\sum_t w_t^{-}\rho_{p(t)}^{-}\Delta t}
{\sum_t w_t^{-}\Delta t}.
\label{eq:rt_weighted_tariffs}
\end{equation}

We organize the analysis by disturbance family and report both coordination
modes together, since the central result is the difference between them.
Per-scenario outcomes for the two modes are collected in
Tables~\ref{tab:rt_profit_based_reference_summary}
and~\ref{tab:rt_operational_based_target_summary}, and the matched mode
difference (operational-based minus profit-based) in
Table~\ref{tab:rt_profit_operational_target_comparison}; the prose highlights
the governing mechanism rather than restating individual values.

\medskip
\noindent\textit{Delay disturbances.}\quad
Timing shifts raise PTO cost in both modes by compressing the windows available
for charging and export, but the modes recover differently. Profit-based
operation holds sell tariffs low, so the aggregator recoups margin from the
charging side and the fleet retains the unexported energy as elevated terminal
SOC (about 28--31\%). Operational-based operation instead pairs a lower buy
tariff with a substantially higher sell tariff, which contains PTO cost
exposure even when discharge windows are reduced; every matched delay case is
cheaper for the PTO than its profit-based counterpart.

\medskip
\noindent\textit{Energy-consumption disturbances.}\quad
The two energy cases act in opposite directions. Under higher consumption
(\(E+50\)) grid purchases rise and export is suppressed (profit-based) or only
partially preserved (operational-based), and PTO cost increases in both modes.
The lower-consumption case (\(E-50\)) is more revealing. In profit-based mode it
produces a counterintuitive outcome: despite reduced traction demand, PTO cost
stays above the DA reference. This follows from the two-step re-optimization ---
an early update exports the temporary SOC surplus while a later update
repurchases energy to restore feasible battery margins once consumption returns
to nominal --- combined with a low V2G compensation tariff that leaves most of
the export value with the aggregator. Operational-based mode turns the same
disturbance into expanded flexibility: it exports 900~kWh (versus 400~kWh) at a
much higher sell tariff, leaving PTO cost essentially unchanged from the DA
reference while raising aggregator revenue only modestly. The contrast shows
that the coordination mode, not the disturbance, governs who captures the value
released by a favorable deviation.

\medskip
\noindent\textit{Price disturbances.}\quad
Price shocks produce the largest economic spread and the sharpest mode
separation. Under positive shocks the profit-based aggregator raises buy tariffs
aggressively and suppresses export, driving PTO cost to the maximum among price-shock scenarios
(247.89~EUR at \(P+50\)) while maximizing its own revenue (73.70~EUR). The
operational-based aggregator instead holds buy tariffs near their DA level and
preserves full V2G compensation, roughly halving PTO cost at the same
disturbance (131.81~EUR) at the expense of its margin. Negative shocks move both
modes in the same direction --- lower buy tariffs, preserved export, reduced PTO
cost --- so the two modes converge when the market itself relieves PTO exposure.

\medskip
\noindent\textit{Combined disturbances.}\quad
Compound stress exposes a structural, not merely quantitative, difference.
Profit-based operation eliminates V2G export entirely across all four combined
cases, so its economic response is driven solely by charging cost and tariff
margin. Operational-based operation preserves export in three of the four cases
(all except \(C\)-All 20--48), keeping flexibility exchange alive under stress.
The exception, \(C\)-All 20--48, is a relaxed-feasibility outcome in both modes:
late-day disturbances leave insufficient time to restore the 20\% reserve while
meeting remaining service commitments, so the SOC floor drops below bound. The
sequential case \(C\)-Seq is less severe than the simultaneous cases in both
modes, because the system can re-optimize between separated disturbances.

\begin{table*}[t]
\centering
\caption{Profit-based RT scenario summary using SOC-derived grid exchange,
including weighted applied tariffs. Deltas are relative to the profit-based DA
reference.}
\label{tab:rt_profit_based_reference_summary}
\begin{adjustbox}{max width=\textwidth}
\footnotesize
\begin{tabular}{llrrrrrrrrr}
\toprule
\makecell[l]{Scenario} & \makecell[l]{Family} &
\makecell{RT PTO\\(EUR)} & \makecell{$\Delta$ PTO\\(EUR)} &
\makecell{RT agg.\\(EUR)} & \makecell{$\Delta$ agg.\\(EUR)} &
\makecell{Buy\\(kWh)} & \makecell{Sell\\(kWh)} &
\makecell{Buy tariff\\(EUR/kWh)} & \makecell{Sell tariff\\(EUR/kWh)} &
\makecell{Final\\SOC (\%)} \\
\midrule
D-30 beg.   & Delay    & 177.88 & +37.30  & 25.83 & +5.52  & 2000 & 100 & 0.0925 & 0.0708 & 30.84 \\
D+30 beg.   & Delay    & 180.08 & +39.49  & 40.24 & +19.93 & 2000 & 200 & 0.0980 & 0.0794 & 30.33 \\
D-30 end    & Delay    & 169.82 & +29.24  & 29.98 & +9.67  & 2000 & 200 & 0.0925 & 0.0757 & 28.49 \\
D+30 end    & Delay    & 177.88 & +37.30  & 25.83 & +5.52  & 2000 & 100 & 0.0925 & 0.0708 & 29.37 \\
E+50        & Energy   & 224.99 & +84.40  & 28.18 & +7.88  & 2388 & 0   & 0.0942 & --     & 27.32 \\
E-50        & Energy   & 161.67 & +21.08  & 45.94 & +25.63 & 2000 & 400 & 0.0925 & 0.0583 & 34.35 \\
P+25        & Price    & 185.89 & +45.31  & 16.90 & -3.41  & 1600 & 0   & 0.1162 & --     & 25.61 \\
P+50        & Price    & 247.89 & +107.30 & 73.70 & +53.40 & 1700 & 81  & 0.1476 & 0.0377 & 28.70 \\
P-25        & Price    & 120.95 & -19.64  & 38.90 & +18.60 & 1900 & 300 & 0.0760 & 0.0780 & 23.44 \\
P-50        & Price    & 70.17  & -70.41  & 33.65 & +13.35 & 2000 & 400 & 0.0507 & 0.0780 & 22.72 \\
C-Seq       & Combined & 225.63 & +85.04  & 57.50 & +37.20 & 1900 & 0   & 0.1188 & --     & 27.54 \\
C-All 5--48 & Combined & 245.37 & +104.78 & 49.07 & +28.77 & 2400 & 0   & 0.1022 & --     & 28.42 \\
C-All 5--25 & Combined & 257.02 & +116.43 & 58.95 & +38.64 & 2400 & 0   & 0.1071 & --     & 42.29 \\
C-All 20--48& Combined & 250.09 & +109.50 & 39.30 & +19.00 & 2400 & 0   & 0.1042 & --     & 27.48 \\
\bottomrule
\end{tabular}
\end{adjustbox}
\end{table*}

\begin{table*}[t]
\centering
\caption{Operational-based RT scenario summary using SOC-derived grid exchange,
including weighted applied tariffs. Deltas are relative to the
operational-based DA reference.}
\label{tab:rt_operational_based_target_summary}
\begin{adjustbox}{max width=\textwidth}
\footnotesize
\begin{tabular}{llrrrrrrrrr}
\toprule
\makecell[l]{Scenario} & \makecell[l]{Family} &
\makecell{RT PTO\\(EUR)} & \makecell{$\Delta$ PTO\\(EUR)} &
\makecell{RT agg.\\(EUR)} & \makecell{$\Delta$ agg.\\(EUR)} &
\makecell{Buy\\(kWh)} & \makecell{Sell\\(kWh)} &
\makecell{Buy tariff\\(EUR/kWh)} & \makecell{Sell tariff\\(EUR/kWh)} &
\makecell{Final\\SOC (\%)} \\
\midrule
D-30 beg.   & Delay    & 157.51 & +38.60  & 8.91  & +6.52  & 2100 & 200 & 0.0863 & 0.1180 & 30.48 \\
D+30 beg.   & Delay    & 137.03 & +18.12  & 9.91  & +7.52  & 2000 & 300 & 0.0858 & 0.1155 & 26.53 \\
D-30 end    & Delay    & 152.56 & +33.65  & 13.38 & +10.99 & 2000 & 200 & 0.0879 & 0.1160 & 28.87 \\
D+30 end    & Delay    & 164.16 & +45.25  & 12.77 & +10.38 & 2000 & 100 & 0.0879 & 0.1160 & 29.74 \\
E+50        & Energy   & 211.09 & +92.18  & 17.22 & +14.83 & 2494 & 100 & 0.0877 & 0.0757 & 29.76 \\
E-50        & Energy   & 118.67 & -0.24   & 17.85 & +15.46 & 2494 & 900 & 0.0877 & 0.1111 & 31.11 \\
P+25        & Price    & 130.85 & +11.94  & 3.28  & +0.89  & 1900 & 300 & 0.0856 & 0.1059 & 23.44 \\
P+50        & Price    & 131.81 & +12.90  & 3.95  & +1.56  & 2000 & 400 & 0.0871 & 0.1059 & 22.72 \\
P-25        & Price    & 80.52  & -38.39  & 3.45  & +1.06  & 2000 & 400 & 0.0614 & 0.1059 & 22.72 \\
P-50        & Price    & 39.56  & -79.35  & 3.04  & +0.65  & 2000 & 400 & 0.0410 & 0.1059 & 22.72 \\
C-Seq       & Combined & 170.28 & +51.37  & 2.41  & +0.02  & 2088 & 100 & 0.0871 & 0.1160 & 29.71 \\
C-All 5--48 & Combined & 196.32 & +77.41  & 3.39  & +1.00  & 2600 & 200 & 0.0844 & 0.1153 & 26.97 \\
C-All 5--25 & Combined & 194.53 & +75.62  & 3.38  & +0.99  & 2600 & 200 & 0.0837 & 0.1160 & 40.84 \\
C-All 20--48& Combined & 222.74 & +103.84 & 12.62 & +10.23 & 2400 & 0   & 0.0928 & --     & 27.86 \\
\bottomrule
\end{tabular}
\end{adjustbox}
\end{table*}

\begin{table*}[t]
\centering
\caption{Matched profit-based--operational-based RT comparison using
SOC-derived grid exchange (operational-based minus profit-based).}
\label{tab:rt_profit_operational_target_comparison}
\begin{adjustbox}{max width=\textwidth}
\begin{tabular}{llrrrrr}
\toprule
\makecell[l]{Scenario} & \makecell[l]{Family} &
\makecell{$\Delta$ PTO\\(EUR)} & \makecell{$\Delta$ agg.\\(EUR)} &
\makecell{$\Delta$ buy\\(kWh)} & \makecell{$\Delta$ sell\\(kWh)} &
\makecell{$\Delta$ final\\SOC\\(pp)} \\
\midrule
D-30 beg.   & Delay    & -20.37  & -16.91 & +100 & +100 & -0.36 \\
D+30 beg.   & Delay    & -43.05  & -30.33 & 0    & +100 & -3.81 \\
D-30 end    & Delay    & -17.26  & -16.60 & 0    & 0    & +0.37 \\
D+30 end    & Delay    & -13.72  & -13.06 & 0    & 0    & +0.37 \\
E+50        & Energy   & -13.90  & -10.96 & +106 & +100 & +2.44 \\
E-50        & Energy   & -43.00  & -28.09 & +494 & +500 & -3.24 \\
P+25        & Price    & -55.05  & -13.62 & +300 & +300 & -2.17 \\
P+50        & Price    & -116.07 & -69.75 & +300 & +319 & -5.98 \\
P-25        & Price    & -40.43  & -35.46 & +100 & +100 & -0.72 \\
P-50        & Price    & -30.61  & -30.61 & 0    & 0    & 0.00  \\
C-Seq       & Combined & -55.35  & -55.09 & +188 & +100 & +2.18 \\
C-All 5--48 & Combined & -49.05  & -45.69 & +200 & +200 & -1.45 \\
C-All 5--25 & Combined & -62.48  & -55.57 & +200 & +200 & -1.45 \\
C-All 20--48& Combined & -27.35  & -26.68 & 0    & 0    & +0.37 \\
\bottomrule
\end{tabular}
\end{adjustbox}
\end{table*}

\begin{figure*}[t]
    \centering
    \begin{subfigure}[t]{0.49\textwidth}
        \centering
        \includegraphics[width=\linewidth]{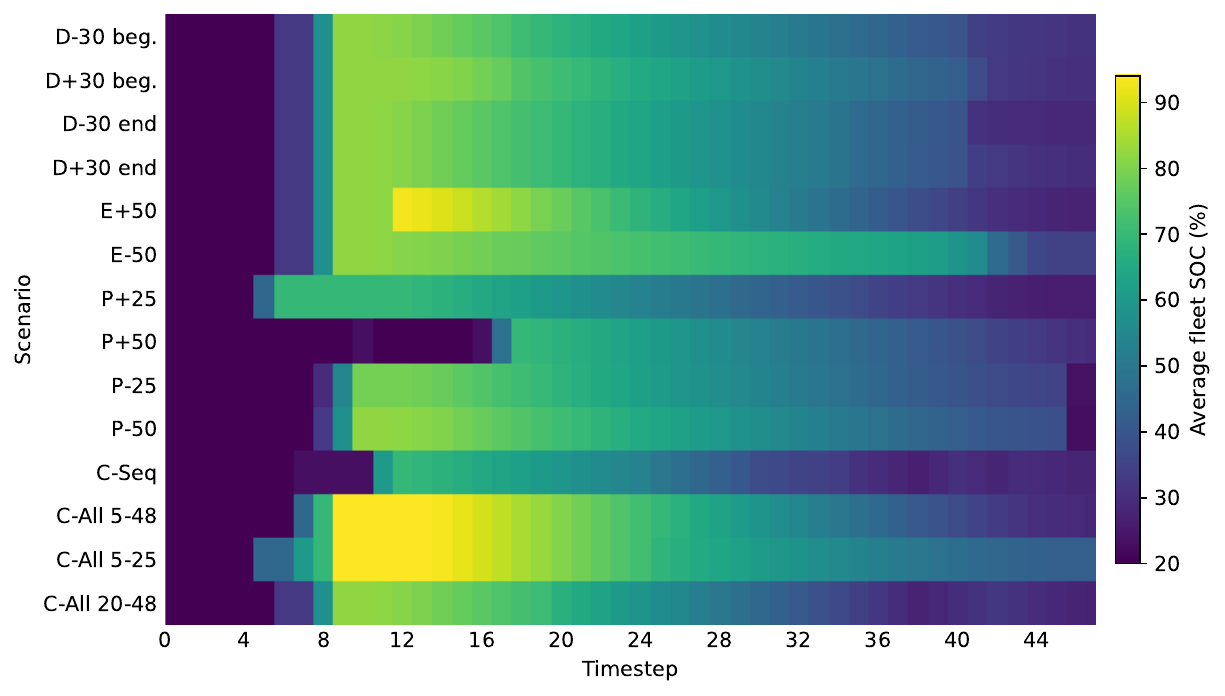}
        \caption{Average fleet SOC.}
        \label{fig:rt_profit_based_soc_heatmap}
    \end{subfigure}
    \hfill
    \begin{subfigure}[t]{0.49\textwidth}
        \centering
        \includegraphics[width=\linewidth]{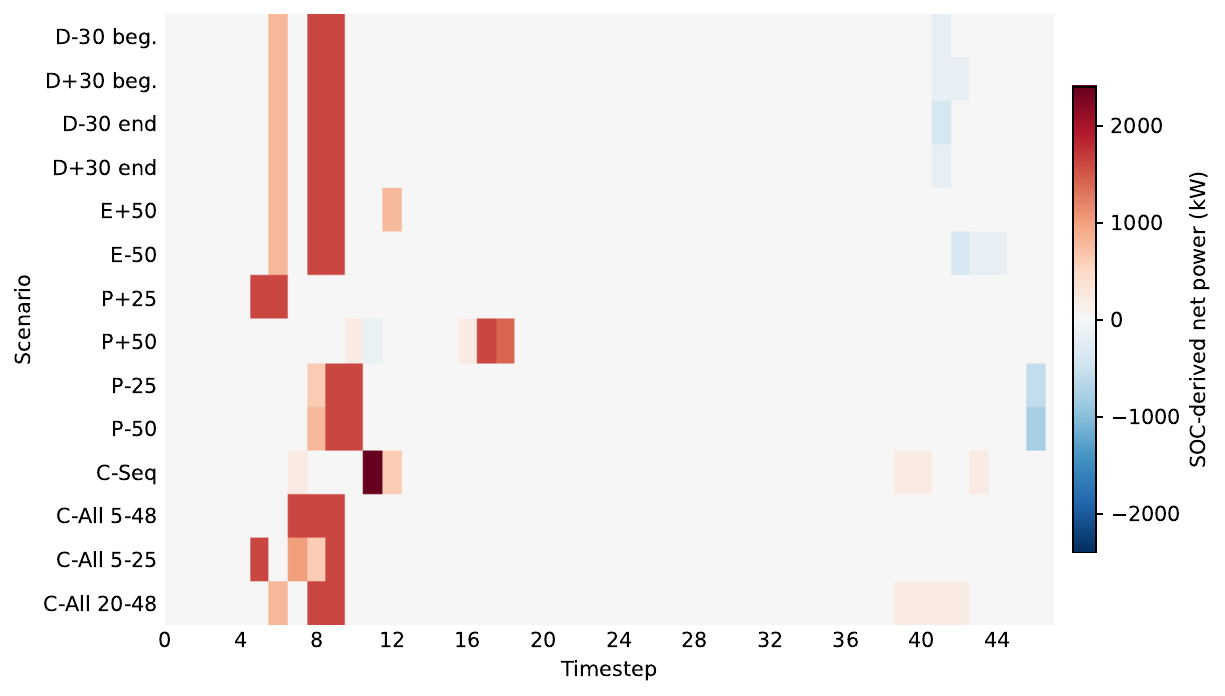}
        \caption{SOC-derived average net power.}
        \label{fig:rt_profit_based_power_heatmap}
    \end{subfigure}
    \caption{Profit-based RT operating heatmaps. Panel~(a) reports average fleet
    SOC. Panel~(b) reports average net power inferred from SOC variation.}
    \label{fig:rt_profit_based_operation_heatmaps}
\end{figure*}

\begin{figure*}[t]
    \centering
    \begin{subfigure}[t]{0.49\textwidth}
        \centering
        \includegraphics[width=\linewidth]{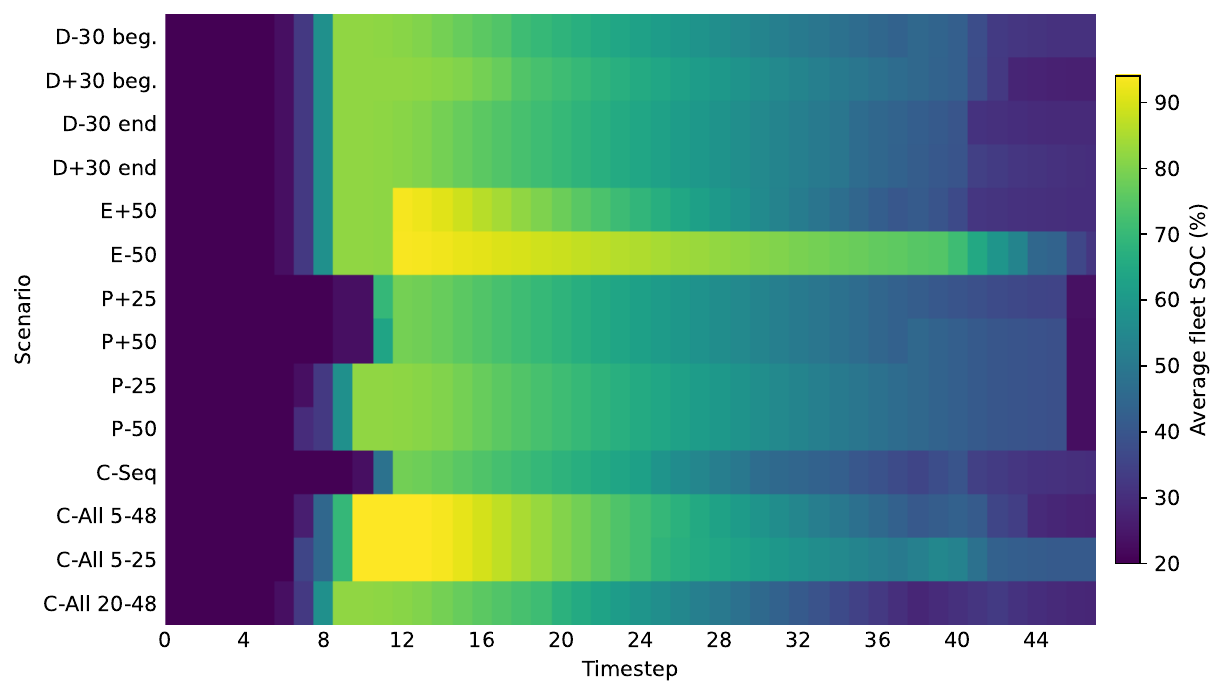}
        \caption{Average fleet SOC.}
        \label{fig:rt_operational_based_soc_heatmap}
    \end{subfigure}
    \hfill
    \begin{subfigure}[t]{0.49\textwidth}
        \centering
        \includegraphics[width=\linewidth]{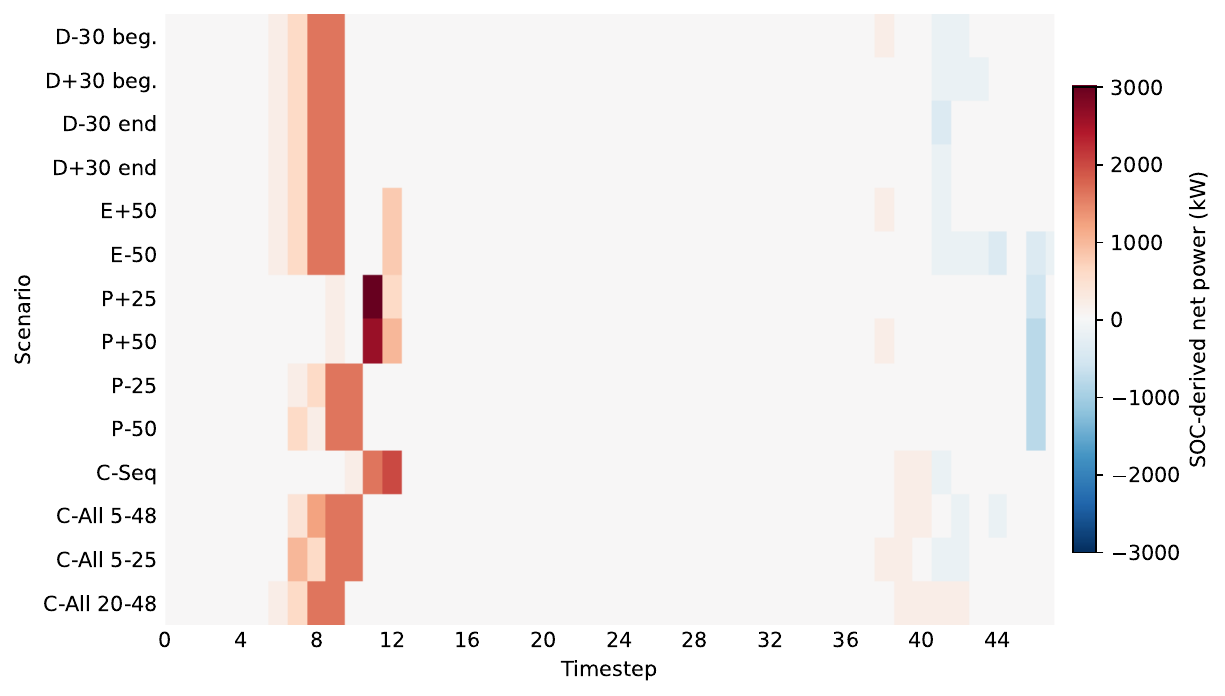}
        \caption{SOC-derived average net power.}
        \label{fig:rt_operational_based_power_heatmap}
    \end{subfigure}
    \caption{Operational-based RT operating heatmaps. Panel~(a) reports average
    fleet SOC. Panel~(b) reports average net power inferred from SOC variation.}
    \label{fig:rt_operational_based_operation_heatmaps}
\end{figure*}

The physical trajectories confirm these economic outcomes.
Figures~\ref{fig:rt_profit_based_operation_heatmaps}
and~\ref{fig:rt_operational_based_operation_heatmaps} show that
operational-based operation sustains larger terminal reserves and broader
export, most visibly in \(E-50\) and the price case, whereas
profit-based operation concentrates charging and curtails discharge,
particularly under combined stress.

\begin{figure*}[t]
    \centering
    \begin{subfigure}[t]{0.49\textwidth}
        \centering
        \includegraphics[width=\linewidth]{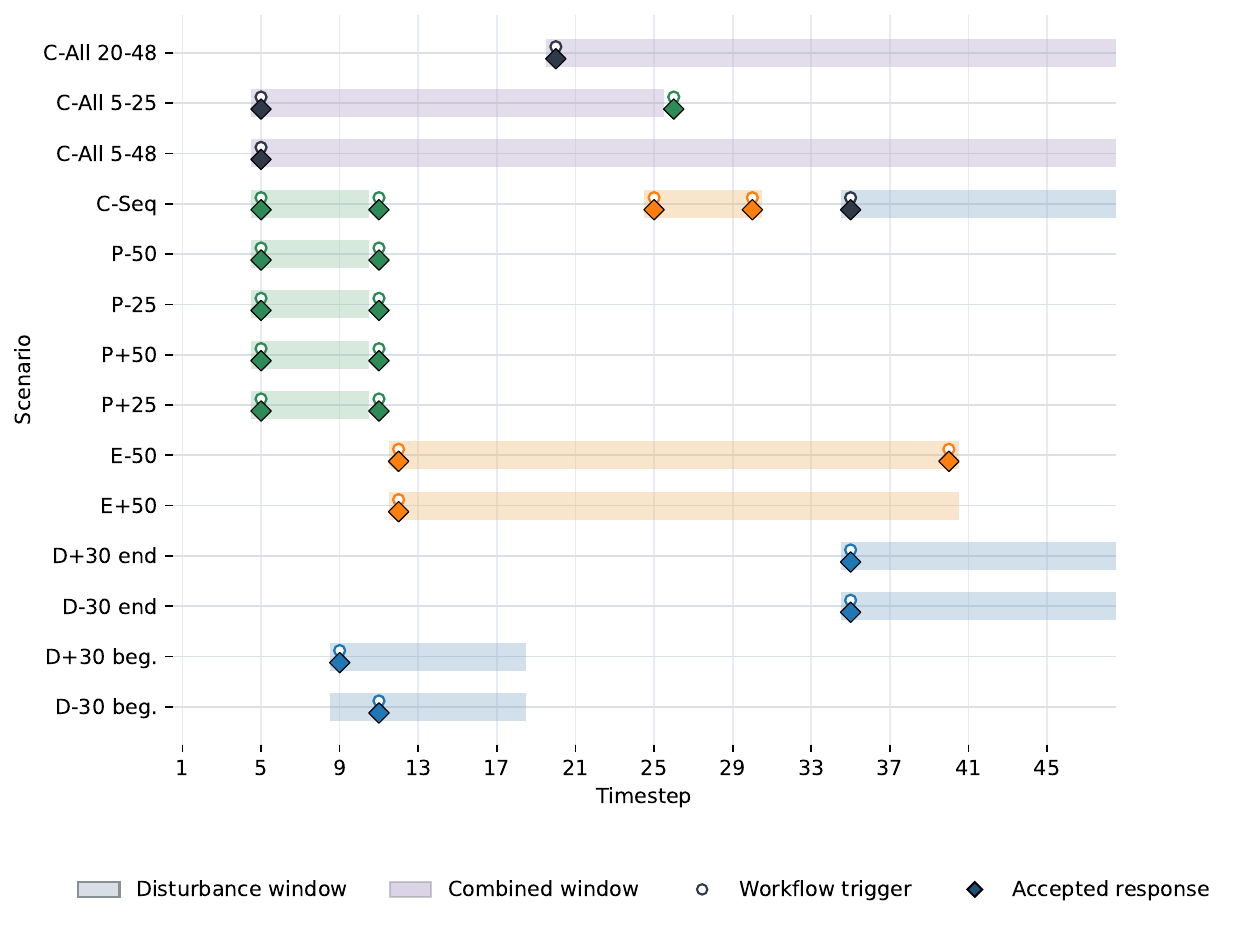}
        \caption{Profit-based.}
    \end{subfigure}
    \hfill
    \begin{subfigure}[t]{0.49\textwidth}
        \centering
        \includegraphics[width=\linewidth]{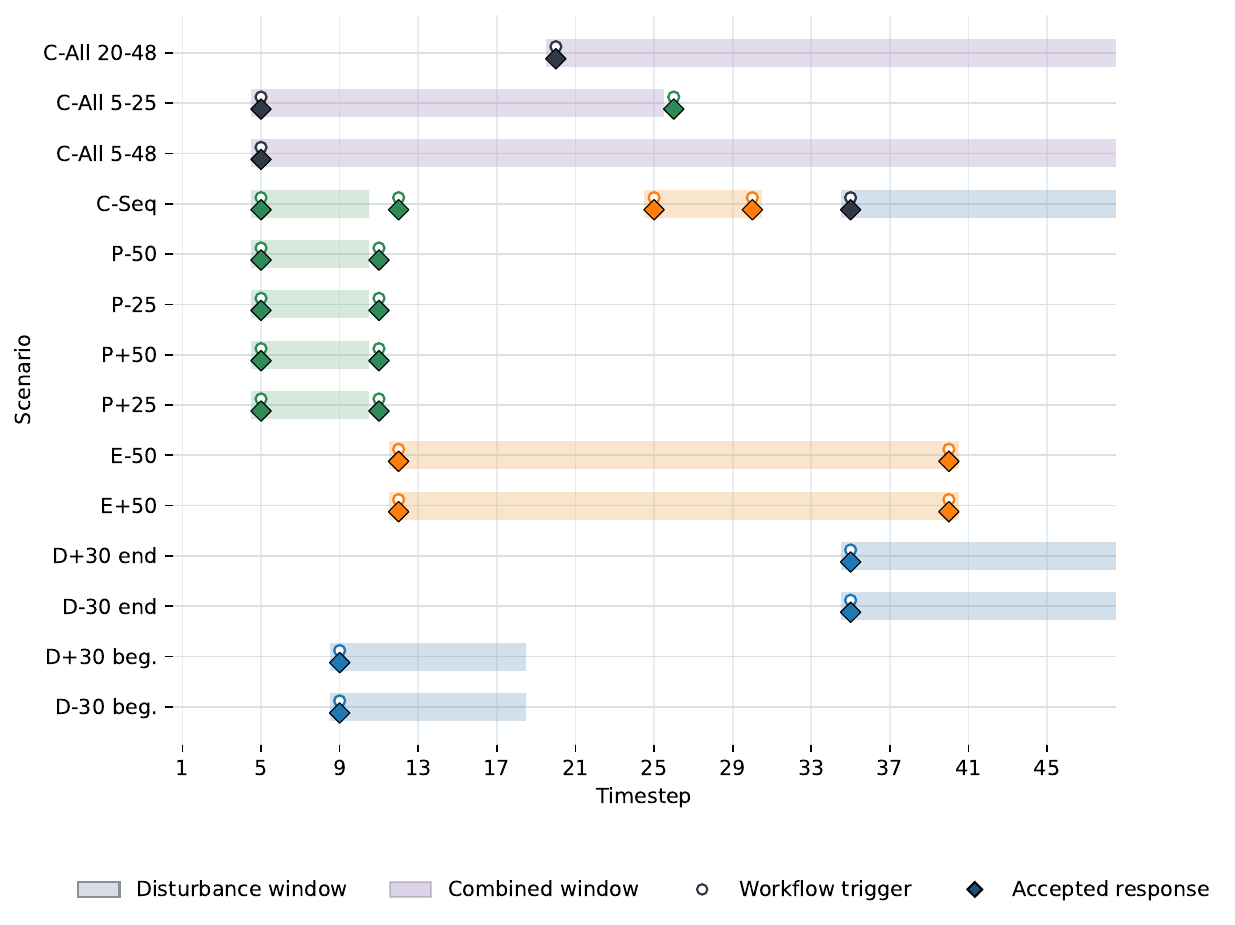}
        \caption{Operational-based.}
    \end{subfigure}
    \caption{RT trigger timelines. Accepted re-optimizations concentrate around
    disturbance-relevant windows in both coordination modes.}
    \label{fig:rt_reoptimization_timelines}
\end{figure*}

In both modes the workflow intervenes selectively rather than continuously:
accepted re-optimizations cluster around disturbance-relevant windows
(Figure~\ref{fig:rt_reoptimization_timelines}), so each update remains traceable
to a specific event and its resulting SOC and tariff trajectory.

\medskip
\noindent\textit{Mode comparison.}\quad
Across all fourteen matched scenarios, operational-based operation reduces PTO
cost without exception, with the largest reductions where profit-based tariffs
are most aggressive (\(P+50\) and the combined cases). This advantage is
mirrored by a fall in aggregator revenue in all fourteen cases. The \(E-50\)
scenario shows why revenue cannot be read from export volume alone:
operational-based mode sells 500~kWh more, yet earns less, because the higher
compensation paid to the PTO outweighs the larger volume. The results therefore
identify no universally dominant mode but a reconciliation frontier between
grid-facing value capture and fleet-facing participation: operational-based
control suits PTO cost containment, service robustness, and contract
acceptability, while profit-based control applies when aggregator revenue is the
objective and the PTO has explicitly accepted the higher tariff exposure. 

Beyond characterizing this frontier, the matched comparison quantifies the economic exposure a PTO faces when the aggregator is profit-oriented and unconstrained. The gap between modes is not marginal: under the (P+50) price shock the PTO pays 116.07~EUR more per day under profit-based coordination than under operational-based coordination, and double-digit daily gaps persist across the combined-disturbance cases. Because the agentic layer can shift between these modes through prompt configuration alone, with no change to the optimizer, the contract, or any externally visible system parameter, this exposure is neither observable to the PTO nor auditable by a regulator under current market arrangements. The same mechanism that lets a cooperative aggregator protect the PTO also lets a profit-oriented one extract value from it, and nothing in the technical architecture distinguishes the two from the outside. This is the empirical basis for the policy argument developed in Section~\ref{sec:policy_implications}: the aggregator domain requires explicit regulatory structure, because the operational-based outcome cannot be assumed to arise on its own.

\subsection{Operational interpretation and decision-support implications}
\label{sec:operational_interpretation}

The results lead to three main operational conclusions that extend beyond the specific scenarios evaluated. (i) DA optimization and V2G pricing are separable decisions with distinct risk profiles. Optimized no-V2G charging reduces PTO cost by 40.2\% relative to dumb charging, showing that fleet operators can capture substantial savings before entering bidirectional flexibility markets. V2G adds further value, but its practical attractiveness depends on how the coordination mode distributes cost savings, export revenue, and terminal reserve between the PTO and the aggregator. (ii) Operational-based aggregation is the most PTO-compatible V2G policy in the evaluated cases. Within the selected FS+CoT configuration used for the main DA comparison, it produces the lowest PTO cost among the four main DA strategies, lowers cost in all fourteen matched RT scenarios, and offers lower buy tariffs with higher sell-back compensation. (iii) Operational-based aggregation is not economically irrational from the aggregator's perspective; it reflects a scale-oriented revenue model in which lower margins per fleet support broader participation, larger flexibility volumes, and more durable contracts. The \(E-50\) case illustrates this trade-off: the operational-based plan exports 500~kWh more than the profit-based plan, while higher sell compensation reduces per-fleet aggregator revenue and raises terminal reserve.

These conclusions also imply a disturbance-dependent acceptance logic for the PTO when an RT re-optimization is proposed. For delay disturbances, the PTO should accept an update when the delay threatens terminal reserve, charger access, or service continuity, rather than only when it improves aggregator revenue. For energy-consumption disturbances, a cost-increasing update can still be acceptable when it restores feasibility or preserves reserve, whereas a lower-consumption update should be checked for unnecessary battery cycling before acceptance. For price disturbances, which affect tariff exposure without disrupting service, a price-only re-optimization should be accepted only when it lowers the PTO bill or follows a pre-agreed market-sharing rule. Across all three families, the acceptance criterion is the PTO's own operational and cost exposure, not the aggregator's revenue, which reinforces the case for the coordination-mode and transparency safeguards discussed in Section~\ref{sec:policy_implications}.

\subsection{Prompt-sensitivity analysis of the Pricing Agent}
\label{sec:prompt_sensitivity}
The preceding DA and RT analyses evaluate the complete agentic optimization workflow. To isolate the behavior of the agentic layer itself, an additional DA experiment was conducted on the Pricing Agent under four prompt paradigms. The optimization model, fleet input data, price profile, and evaluator structure were kept fixed; only the prompt used to generate the DA buy and sell multipliers was changed. The four prompt scenarios are zero-shot (ZS), chain-of-thought prompting (CoT), few-shot prompting (FS), and few-shot plus chain-of-thought prompting (FS+CoT). Here, ZS provides the role, task, economic context, and output requirements without worked examples; CoT adds an explicit reasoning scaffold; FS adds example-guided pricing behavior; and FS+CoT combines examples with structured reasoning. Since the RT experiments use zero-shot role prompting with constrained output formatting, this sensitivity analysis focuses on the DA Pricing Agent, where the prompt variants were explicitly tested.

Table~\ref{tab:prompt_sensitivity_results} reports the accepted DA outcomes from the prompt-experiment workbooks using the same SOC-derived accounting as Table~\ref{tab:da_all_results}. The FS+CoT rows therefore correspond to the accepted profit-based and operational-based DA files used in the main DA analysis. \ref{app:compact_prompt_templates} provides compact versions of the tested prompt templates.

Under profit-based behavior, prompt design strongly affects the agent's ability to express the intended aggregator objective. Aggregator revenue increases from 6.66 EUR/day under ZS to 10.61 EUR/day under FS, 15.96 EUR/day under CoT, and 20.30 EUR/day under FS+CoT. This ordering indicates that examples help the agent identify useful tariff patterns, while the reasoning scaffold helps connect tariff spread, V2G volume, and aggregator margin. The operational-based results show a different pattern: PTO costs are tightly clustered, spanning only 0.71 EUR/day between the best and worst cases. FS gives the lowest PTO cost, 118.51 EUR/day, while FS+CoT gives the lowest aggregator revenue, 2.39 EUR/day, and remains close to the best PTO outcome. CoT alone performs worst for operational-based pricing, increasing both PTO cost and aggregator revenue. This suggests that reasoning without examples can lead the agent to over-emphasize economically plausible tariff spread rather than the PTO-facing objective, while examples provide the main alignment signal in the operational-based case.

These experiments support the interpretation that the proposed framework is not only an optimization wrapper, but an agentic decision system whose behavior depends on how economic objectives are communicated to the agent. The optimizer enforces feasibility once tariffs are provided, but the prompt controls the tariff hypothesis evaluated by the optimizer. Prompting is therefore part of the experimental design, and prompt paradigms should be reported as experimental scenarios and treated as a reproducibility variable in agentic energy-management studies.

\begin{table}[t]
	\centering
	\caption{Pricing Agent prompt-sensitivity results.}
	\label{tab:prompt_sensitivity_results}
	\small
	\setlength{\tabcolsep}{4pt}
	\renewcommand{\arraystretch}{1.05}
	\begin{threeparttable}
		\begin{tabularx}{\columnwidth}{@{}
				l
				l
				>{\centering\arraybackslash}X
				>{\centering\arraybackslash}X
				c
				@{}}
			\toprule
			Mode 
			& Prompt\tnote{a}
			& \makecell[c]{PTO cost\\(EUR/day)}
			& \makecell[c]{Aggregator\\ revenue\\(EUR/day)}
			& \makecell[c]{rank\tnote{b}} \\
			\midrule
			PB & ZS     & 137.13 &  6.66 & 4 \\
			PB & FS     & 137.31 & 10.61 & 3 \\
			PB & CoT    & 135.23 & 15.96 & 2 \\
			PB & FS+CoT & 140.59 & 20.30 & 1 \\
			OB & ZS     & 118.83 &  2.75 & 2 \\
			OB & FS     & 118.51 &  2.43 & 1 \\
			OB & CoT    & 119.22 &  3.15 & 4 \\
			OB & FS+CoT & 118.91 &  2.39 & 3 \\
			\bottomrule
		\end{tabularx}
		\begin{tablenotes}
			\footnotesize
			\item[a] PB: profit-based; OB: operational-based; ZS: zero-shot; FS: few-shot; CoT: chain-of-thought; FS+CoT: few-shot + CoT.
			\item[b] Rank 1 = best under mode objective: max aggregator revenue (PB) or min PTO cost (OB).
		\end{tablenotes}
	\end{threeparttable}
\end{table}
\subsection{Policy Implications}
\label{sec:policy_implications}

The preceding results are operational, but their strongest implication is regulatory: the aggregator domain cannot be left to unconstrained market behavior if electric bus fleets are to participate in V2G markets on sustainable terms. Three findings established above support this conclusion.

First, the coordination mode, not the disturbance, governs how flexibility value is split between the aggregator and the PTO, and the gap is large --- up to 116.07~EUR/day under the \(P+50\) shock. A profit-maximizing aggregator that raises PTO cost is not behaving anomalously; it is doing exactly what an unconstrained revenue objective implies. Absent a rule that bounds tariff margins or mandates value-sharing, the PTO, a public entity operating on public budgets, absorbs this exposure directly.

Second, for an agentic aggregator this pricing aggressiveness is set by prompt configuration alone, which is invisible from outside the system (Section~\ref{sec:prompt_sensitivity}). This is a qualitatively different regulatory problem from classical aggregator design: a rule-based pricing scheme can be audited from its parameters, whereas an LLM-based one can be re-tuned through prompt edits that leave no structural trace, and that disclosure rules written for conventional aggregators would not capture.

Third, the PTO-protective operational-based outcome is not the aggregator's default incentive, since it reduces per-fleet revenue in every matched scenario. A profit-oriented operator has no self-interested reason to adopt it unless its revenue model is explicitly scale-based or its margins are externally constrained. The PTO-favorable equilibrium must therefore be induced, not assumed.

These findings motivate three regulatory measures for V2G markets involving public fleets: (i) bounds on aggregator tariff margins, or mandated value-sharing rules, so that PTO cost exposure cannot exceed a defined level regardless of the aggregator's objective; (ii) transparency requirements covering the coordination mode and, for agentic systems, the prompt configuration and constrained-output specification that govern pricing behavior, so that the chosen objective is auditable; and (iii) reporting standards requiring disclosure of realized tariff vectors, trigger decisions, and value allocation, extending existing market-conduct oversight to the agentic setting.

A related implication concerns the role of human stakeholders in governing agentic aggregators. Human oversight should operate primarily at the governance level rather than through manual approval of every charging or V2G action. In practice, human control should be placed at the level of objectives, constraints, audit, and intervention authority. PTOs, aggregators, and regulators should define the admissible coordination mode, tariff-margin bounds, SOC and service-reliability safeguards, and conditions under which the system may re-optimize or switch pricing behavior. Human decision-makers should also have access to interpretable records of prompt configuration, tariff vectors, trigger decisions, accepted schedules, and value allocation, so that agentic decisions can be reviewed and overridden when they conflict with public-service priorities. In this framing, the agentic aggregator remains an automated operational layer, but its economic posture and safety envelope remain human-governed.

\section{Conclusions}
\label{sec:conclusion}

This paper proposed a multi-agentic framework for electric bus fleet-grid coordination under DA and RT operation. The aggregator is framed as a reconciliation entity that converts grid-side flexibility signals into fleet-feasible charging and V2G decisions while allocating value between the aggregator and the PTO. The framework combines an optimization-based charging model with three supervisory agents: a Trigger Agent that determines whether updated conditions justify intervention, a Pricing Agent that revises tariff guidance, and an Evaluator Agent that accepts or rejects the resulting operating plan. In this architecture, the DA layer defines the nominal charging and V2G reference, while the RT layer receives updated information in 30-minute blocks and revises the plan only when disturbances become operationally or economically relevant. The contribution is therefore a structured decision pipeline that separates disturbance detection, tariff adaptation, and schedule acceptance while preserving the transparency of an optimization-based scheduler.

The DA results establish the economic baseline. Optimized no-V2G charging reduces PTO cost by 40.2\% relative to dumb charging, showing that substantial fleet savings are available even before bidirectional flexibility is considered. The V2G cases then show that the aggregator coordination mode determines how the additional value is distributed. The profit-based mode produces the highest DA aggregator revenue, 20.30 EUR/day, by maintaining a larger spread between PTO buy and sell-back tariffs. Within the selected FS+CoT configuration used for the main DA comparison, the operational-based mode produces the lowest PTO cost among the four main DA strategies, 118.91 EUR/day, by lowering the weighted buy tariff and increasing the sell-back tariff. The prompt-sensitivity analysis further shows that the FS operational-based prompt reduces this cost to 118.51 EUR/day. Thus, the aggregator objective is not a neutral modeling detail; it directly controls the allocation of cost savings, V2G revenue, and terminal reserve. The RT results strengthen this conclusion under disturbance conditions. Across the fourteen matched RT scenarios, operational-based operation lowers PTO cost in every case relative to profit-based operation. The largest PTO reductions occur under \(P+50\) (-116.07 EUR), \(C\)-All 5--25 (-62.48 EUR), \(C\)-Seq (-55.35 EUR), and \(P+25\) (-55.05 EUR), showing that operational-based pricing is especially important when positive price shocks, early service disruptions, or compound disturbances would otherwise expose the PTO to high tariff margins or reduced V2G value. The PTO benefit is not free from the aggregator's perspective: operational-based operation reduces aggregator revenue in all fourteen matched scenarios. In \(E-50\), the lower energy-consumption disturbance creates enough additional stored energy for operational-based dispatch to sell 900 kWh, but the higher compensation paid to the PTO keeps aggregator revenue below the profit-based case. The RT results therefore show that aggregator revenue is governed by both tariff margin and available flexibility volume, not by tariff margin alone. The prompt-sensitivity experiment confirms that the agentic layer is itself an experimental object, not only an interface to the optimizer. For profit-based pricing, the objective ranking is FS+CoT, CoT, FS, then ZS, with SOC-derived aggregator revenue increasing from 6.66 EUR/day under ZS to 20.30 EUR/day under FS+CoT. For operational-based pricing, examples are more important than reasoning alone: FS gives the lowest SOC-derived PTO cost, 118.51 EUR/day, while FS+CoT gives the lowest aggregator revenue and remains within 0.40 EUR/day of the best PTO-cost result. CoT alone performs worst because it raises PTO cost and aggregator revenue relative to the example-guided prompts. These results show that prompt design affects whether the Pricing Agent expresses the intended economic behavior. Future agentic energy-management studies should therefore report prompt paradigms and constrained-output assumptions alongside optimization inputs and results.

From the fleet perspective, relying only on the no-V2G optimum can be economically beneficial, but it leaves bidirectional flexibility value unused. If V2G is adopted, the operational-based coordination mode is the PTO-facing policy that consistently makes sense in the evaluated scenarios because it lowers PTO cost in all matched RT cases and offers more favorable tariff exposure. From the aggregator perspective, operational-based operation is more compatible with durable fleet participation and contract acceptance, but it generally reduces revenue per fleet. Large aggregator revenues should therefore come primarily from increasing the number, size, and diversity of aggregated fleets, and from mobilizing larger volumes of flexibility, rather than from extracting high margins from one PTO fleet. This is the central operational implication of the reconciliation framing: scalable grid flexibility from electric bus fleets depends on participation terms that remain acceptable to the PTO.

Beyond these operational lessons, the results carry a regulatory implication that is, in our view, the broader contribution of the study. Because the coordination mode alone moves PTO cost by up to 116.07~EUR/day in the evaluated cases, and because an agentic aggregator can switch between modes through prompt configuration that leaves no externally visible trace, the PTO-favorable outcome cannot be assumed to arise on its own and cannot be verified by a counterparty under current market arrangements. The aggregator domain therefore warrants explicit regulatory structure when public transport fleets are involved: bounds on tariff margins or mandated value-sharing to cap PTO cost exposure, transparency requirements covering the coordination mode and the prompt and constrained-output specifications that govern agentic pricing, and reporting standards that extend market-conduct oversight to realized tariff vectors, trigger decisions, and value allocation. As agentic AI is adopted in energy-market intermediation, the framework proposed here doubles as a diagnostic instrument that makes the mode- and prompt-dependent allocation of value explicit, and thereby identifies where such oversight is most needed.


The results should be interpreted within the limits of the present study. The RT evaluation uses controlled isolated and combined disturbance cases, and it compares accepted RT responses against the DA coordination references without adding no-action, rule-based trigger, or always-reoptimize baselines. The experiments also rely on a single depot configuration, charger setting, price profile, and initial SOC structure. The policy implications drawn above follow from this single-setting evaluation and should be read as evidence-based motivation for regulatory attention rather than as calibrated threshold values. Future work should extend the evaluation to larger and more heterogeneous depots, different charger-to-bus ratios, alternative market conditions, richer combined-disturbance designs, stochastic repeated runs, prompt-sensitivity analysis, solver-runtime measurement, and formal service-reliability metrics. Furthermore, Cybersecurity issues of agentic systems, as mentioned in \cite{eslami2026stability} and \cite{eslami2025security} will also be investigated. The main methodological lesson is that RT fleet-grid coordination should report not only aggregate cost and revenue, but also the trigger timing, accepted tariff vectors, SOC trajectories, and V2G exchange pathways through which disturbances reshape operational outcomes.



\section*{Declaration of Competing Interest}
The authors declare that they have no known competing financial interests or personal relationships that could have influenced the work reported in this paper.

\section*{Data Availability Statement}
The data and source code that support the findings of this study are openly available in the repository at \url{https://github.com/AlienEslami/Agentic-Aggregator}.

\section*{Acknowledgment}
This project was made possible through financial support from the Government of Canada and the Fonds de recherche du Québec – Nature et technologies (FRQNT). Funding from the Government of Canada's Environmental Damages Fund was provided under its Climate Action and Awareness Fund. At the same time, the FRQNT contributed through the Programme de recherche en partenariat - Réduction des GES - Mobilité Durable.


\appendix

\renewcommand{\thesection}{Appendix~\Alph{section}}
\renewcommand{\thesubsection}{\Alph{section}.\arabic{subsection}}

\renewcommand{\thefigure}{\Alph{section}.\arabic{figure}}
\renewcommand{\thetable}{\Alph{section}.\arabic{table}}
\renewcommand{\theequation}{\Alph{section}.\arabic{equation}}

\makeatletter
\@addtoreset{figure}{section}
\@addtoreset{table}{section}
\@addtoreset{equation}{section}
\makeatother

\section{Nomenclature}

\addcontentsline{toc}{section}{Nomenclature}
\footnotesize\RaggedRight

\nomgroup{Indices}
\nomentry{$p$}{Tariff-period index.}
\nomentry{$t$}{Time-interval index.}
\nomentry{$k$}{Bus index.}
\nomentry{$n$}{Charger index.}
\nomentry{$i$}{Route or service-block index.}
\nomentry{$\tau$}{Current operating interval in the RT workflow.}
\nomentry{$r$}{Rerun index in the DA or RT workflow.}
\nomentry{$d$}{Operating-day index.}

\nomgroup{Sets}
\nomentry{$\mathcal{P}$}{Set of tariff periods.}
\nomentry{$\mathcal{T}$}{Set of time intervals in the DA optimization horizon.}
\nomentry{$\mathcal{T}_{\tau}^{\mathrm{RT}}$}{Set of RT optimization intervals from $\tau$ to the end of the operating day.}
\nomentry{$\mathcal{K}$}{Set of electric buses.}
\nomentry{$\mathcal{N}$}{Set of chargers.}
\nomentry{$\mathcal{I}$}{Set of routes or service blocks.}

\nomgroup{Time and workflow parameters}
\nomentry{$T$}{Final interval of the operating horizon.}
\nomentry{$\Delta t$}{Duration of one time interval.}
\nomentry{$p(t)$}{Tariff period associated with time interval $t$.}
\nomentry{$R_{\max}$}{Maximum number of allowed optimization reruns.}

\nomgroup{Tariff and market parameters}
\nomentry{$\pi_t^{g}$}{Grid energy price at time interval $t$.}
\nomentry{$\pi_t^{g,\mathrm{obs}}$}{Observed or updated grid energy price used in RT at interval $t$.}
\nomentry{$\boldsymbol{\pi}_{\tau:T}^{g,\mathrm{obs}}$}{Updated grid-price profile used from RT interval $\tau$ to $T$.}
\nomentry{$\bar{\pi}_{p}^{g}$}{Average grid energy price in tariff period $p$.}
\nomentry{$\rho_p^{+}$}{Tariff charged by the aggregator to the PTO for charging energy in tariff period $p$.}
\nomentry{$\rho_p^{-}$}{Tariff paid by the aggregator to the PTO for V2G energy in tariff period $p$.}
\nomentry{$\boldsymbol{\rho}^{+}$}{Vector of charging-energy tariffs.}
\nomentry{$\boldsymbol{\rho}^{-}$}{Vector of V2G compensation tariffs.}
\nomentry{$\alpha_p^{+}$}{Charging-energy tariff multiplier proposed by the Pricing Agent in tariff period $p$.}
\nomentry{$\alpha_p^{-}$}{V2G compensation tariff multiplier proposed by the Pricing Agent in tariff period $p$.}
\nomentry{$\underline{\alpha}_{m}^{+},\overline{\alpha}_{m}^{+}$}{Lower and upper bounds on charging-energy tariff multipliers under mode $m$.}
\nomentry{$\underline{\alpha}_{m}^{-},\overline{\alpha}_{m}^{-}$}{Lower and upper bounds on V2G compensation tariff multipliers under mode $m$.}

\nomgroup{Fleet, battery, and charger parameters}
\nomentry{$P_n^{\mathrm{ch}}$}{Charging power rating of charger $n$.}
\nomentry{$P_n^{\mathrm{dis}}$}{V2G discharging power rating of charger $n$.}
\nomentry{$\eta_n^{\mathrm{ch}}$}{Charging efficiency of charger $n$.}
\nomentry{$\eta_n^{\mathrm{dis}}$}{Discharging efficiency of charger $n$.}
\nomentry{$E_k^{\min}$}{Minimum allowable stored energy of bus $k$.}
\nomentry{$E_k^{\max}$}{Maximum allowable stored energy of bus $k$.}
\nomentry{$E_k^{0}$}{Initial stored energy of bus $k$ in the DA optimization.}
\nomentry{$E_k^{\mathrm{end}}$}{Required end-of-day stored energy of bus $k$.}
\nomentry{$\xi_i$}{Route-level energy requirement for route or service block $i$.}
\nomentry{$\xi_i^{\mathrm{obs}}$}{Observed route-level energy requirement used in RT operation.}
\nomentry{$\xi_i^{\mathrm{RT}}$}{Updated RT route-level energy requirement.}
\nomentry{$\boldsymbol{\xi}^{\mathrm{obs}}$}{Vector of observed route-level energy requirements used in RT operation.}
\nomentry{$\overline{W}$}{Maximum admissible aggregate depot power exchange.}


\nomgroup{Decision variables}
\nomentry{$x_{k,n,t}$}{Binary variable equal to 1 if bus $k$ charges at charger $n$ during interval $t$.}
\nomentry{$y_{k,n,t}$}{Binary variable equal to 1 if bus $k$ discharges through charger $n$ during interval $t$.}
\nomentry{$b_{k,i,t}$}{Binary variable equal to 1 if bus $k$ is assigned to route or service block $i$ during interval $t$.}
\nomentry{$\bar{b}_{k,i,t}^{\mathrm{active}}$}{Indicator fixing route assignments already underway in RT operation.}
\nomentry{$e_{k,t}$}{Stored energy of bus $k$ at time interval $t$.}
\nomentry{$w_t^{+}$}{Aggregate charging power purchased by the PTO from the aggregator at interval $t$.}
\nomentry{$w_t^{-}$}{Aggregate V2G discharging power sold by the PTO to the aggregator at interval $t$.}
\nomentry{$\mathbf{s}$}{PTO scheduling decision vector.}
\nomentry{$\mathbf{s}^{*}$}{Optimal or accepted PTO schedule.}
\nomentry{$\mathbf{s}_{d,r}^{*}$}{Candidate DA schedule at operating day $d$ and rerun $r$.}
\nomentry{$\mathbf{s}_{\mathrm{DA}}^{*}$}{Accepted DA reference schedule.}
\nomentry{$\mathbf{s}_{\tau-1}^{*}$}{Last accepted schedule before RT interval $\tau$.}
\nomentry{$\mathbf{s}_{\tau}^{*}$}{Accepted schedule at RT interval $\tau$.}
\nomentry{$\mathbf{s}_{\tau,r}^{*}$}{Candidate RT schedule at interval $\tau$ and rerun $r$.}
\nomentry{$\mathbf{s}_{\tau}^{\mathrm{RT},*}$}{Accepted RT schedule computed from interval $\tau$ onward.}
\nomentry{$\mathbf{s}_{\mathrm{ref}}$}{Active reference schedule used during RT operation.}

\normalsize\justifying

\section{Optimization Core}
\label{sec:bilevel}

Having established the supervisory agent roles in
Section~\ref{sec:multiagent}, this Appendix presents the mathematical
problem they govern.
The aggregator pricing function is fully assumed by the agentic layer;
what remains here is the PTO scheduling model that the agents
orchestrate, the real-time re-optimization logic that is triggered by
the agents, and the formal agentic supervisory flow that connects the
two.
The PTO scheduling problem is optimization-based throughout, since
route service, battery feasibility, and charger availability must
satisfy explicit physical constraints regardless of how the surrounding
decision process is orchestrated.

\subsection{PTO Scheduling Model: Full Formulation}
\label{sec:pto_optimization_core}

Let $p \in \mathcal{P}$ denote tariff periods, $t \in \mathcal{T}$
time intervals, $k \in \mathcal{K}$ buses, $n \in \mathcal{N}$
chargers, and $i \in \mathcal{I}$ routes.
At each DA or RT invocation, the Pricing Agent supplies two tariff
vectors: $\boldsymbol{\rho}^{+}$, the price charged to the PTO for
charging energy, and $\boldsymbol{\rho}^{-}$, the price paid to the
PTO for V2G energy.
The relationship between these tariff vectors and the agent-proposed
multipliers is defined in Appendix~\ref{sec:agentic_logic}.
Given these tariffs, the PTO selects a schedule $\mathbf{s}^{*}$ that
minimizes its net energy-trading cost:
\begin{equation}
\mathbf{s}^{*}
\in
\arg\min_{\mathbf{s}\in\mathcal{S}(\boldsymbol{\rho}^{+},\boldsymbol{\rho}^{-})}
C^{\mathrm{PTO}}
=
\sum_{t \in \mathcal{T}}
\left[
\rho_{p(t)}^{+} w_t^{+}\Delta t
-
\rho_{p(t)}^{-} w_t^{-}\Delta t
\right],
\label{eq:lower}
\end{equation}
where $\mathcal{S}(\boldsymbol{\rho}^{+},\boldsymbol{\rho}^{-})$ is
the feasible set defined by the operational constraints below.
The objective minimizes the cost of charging energy net of V2G
compensation.

The decision vector includes the route-assignment variables $b_{k,i,t}$,
the charging variables $x_{k,n,t}$, the V2G discharging variables
$y_{k,n,t}$, the stored energy variables $e_{k,t}$, and the aggregate
depot power variables $w_t^{+}$ and $w_t^{-}$.

A bus cannot serve a route and use a charger at the same time.
This operational exclusivity is represented by
Eq.~\eqref{eq:trip_conflict}:
\begin{equation}
\sum_{n \in \mathcal{N}} \left(x_{k,n,t}+y_{k,n,t}\right)
+
\sum_{i \in \mathcal{I}} b_{k,i,t}
\le 1,
\qquad
\forall k\in \mathcal{K},\forall t\in \mathcal{T}.
\label{eq:trip_conflict}
\end{equation}
The first term identifies whether bus $k$ is connected to a charger
for charging or discharging, while the second term identifies whether
it is assigned to a route.
This constraint prevents physically incompatible decisions, such as
charging a bus while it is in service.
The stored energy of each bus evolves according to
Eq.~\eqref{eq:soc}:
\begin{equation}
\begin{aligned}
	\begin{split}
e_{k,t+1}
&=
e_{k,t}
+
\sum_{n \in \mathcal{N}} \eta_n^{\mathrm{ch}} P_n^{\mathrm{ch}} x_{k,n,t}\Delta t
-
\sum_{n \in \mathcal{N}} \frac{1}{\eta_n^{\mathrm{dis}}} P_n^{\mathrm{dis}} y_{k,n,t}\Delta t
\\&-
\sum_{i \in \mathcal{I}}\xi_i b_{k,i,t},
\quad
\forall k\in \mathcal{K},\forall t\in \mathcal{T}.
\end{split}
\end{aligned}
\label{eq:soc}
\end{equation}
The second term represents the energy added through charging,
corrected by the charging efficiency.
The third term represents the energy removed through V2G discharge,
corrected by the discharging efficiency.
The final term represents the traction energy consumed when bus $k$
serves route $i$, where $\xi_i$ is the energy consumption.
Battery feasibility is enforced through Eq.~\eqref{eq:soc_bounds}:
\begin{equation}
	\begin{split}
E_k^{\min} &\le e_{k,t} \le E_k^{\max},
\qquad
e_{k,0} = E_k^{0},\\&
e_{k,T} \ge E_k^{\mathrm{end}},
\qquad
\forall k\in \mathcal{K},\forall t\in \mathcal{T}.
\end{split}
\label{eq:soc_bounds}
\end{equation}
These constraints ensure that each bus remains within its admissible
battery range, starts from the specified initial energy level, and
finishes the horizon with enough energy for subsequent operation.
Charging and discharging are mutually exclusive for each bus--charger
pair as presented in Eq.~\eqref{eq:charger_occupancy}:
\begin{equation}
x_{k,n,t} + y_{k,n,t} \le 1,
\qquad
\forall k \in \mathcal{K},\; \forall n \in \mathcal{N},\; \forall t \in \mathcal{T}.
\label{eq:charger_occupancy}
\end{equation}
Additional charger-capacity constraints ensure that each charger
serves at most one bus in each time interval.
At the depot level, individual charging and discharging decisions are
aggregated as in
Eqs.~\eqref{eq:exchange_power_charge}--\eqref{eq:exchange_power_discharge}:
\begin{equation}
w_t^+
=
\sum_{k \in \mathcal{K}}\sum_{n \in \mathcal{N}}
P_n^{\mathrm{ch}}x_{k,n,t},
\qquad
\forall t\in \mathcal{T},
\label{eq:exchange_power_charge}
\end{equation}
\begin{equation}
w_t^-
=
\sum_{k \in \mathcal{K}}\sum_{n \in \mathcal{N}}
P_n^{\mathrm{dis}}y_{k,n,t},
\qquad
\forall t\in \mathcal{T}.
\label{eq:exchange_power_discharge}
\end{equation}
The depot exchange is bounded by the admissible power limit as in
Eq.~\eqref{eq:power_limits}:
\begin{equation}
0 \le w_t^+ \le \overline{W},
\qquad
0 \le w_t^- \le \overline{W},
\qquad
\forall t\in \mathcal{T}.
\label{eq:power_limits}
\end{equation}

\subsection{Real-Time Re-Optimization Logic}
\label{sec:rt_optimization_logic}

The DA formulation computes a full-day reference schedule before
operation begins.
During operation, this reference may become outdated because the
observed system state can differ from the DA assumptions.
These deviations may include differences between planned and observed
bus SOC, service delays, updated electricity prices, or changes in the
route-level energy required to complete service.
The RT workflow addresses these deviations by rerunning the PTO
optimization model. The RT workflow logic is summarized in Fig.~\ref{fig:rt_logic}.

Let $\tau$ denote the current operating interval.
If the Trigger Agent decides to re-optimize, the optimization horizon
is updated from the current interval to the end of the operating day
as in Eq.~\eqref{eq:rt_horizon}:
\begin{equation}
\mathcal{T}_{\tau}^{\mathrm{RT}}=\{\tau,\tau+1,\ldots,T\}.
\label{eq:rt_horizon}
\end{equation}
The RT optimization does not revise decisions for intervals that have
already occurred.
It uses the observed state at interval $\tau$ as the new initial
condition and optimizes the remaining charging, discharging, and
service decisions from that point onward.
At each RT update, the observed state is represented as in
Eq.~\eqref{eq:rt_state}:
\begin{equation}
\mathbf{x}_{\tau}^{\mathrm{obs}}
=
\left(
\mathbf{e}_{\tau}^{\mathrm{obs}},
\mathbf{d}_{\tau}^{\mathrm{obs}},
\boldsymbol{\pi}_{\tau:T}^{g,\mathrm{obs}},
\boldsymbol{\xi}^{\mathrm{obs}}
\right),
\label{eq:rt_state}
\end{equation}
where $\mathbf{e}_{\tau}^{\mathrm{obs}}$ is the observed bus-energy
vector at the current interval, $\mathbf{d}_{\tau}^{\mathrm{obs}}$
contains the observed service-timing state,
$\boldsymbol{\pi}_{\tau:T}^{g,\mathrm{obs}}$ is the updated
electricity-price profile used for the remaining horizon, and
$\boldsymbol{\xi}^{\mathrm{obs}}$ contains the observed route-level
energy consumption.
Once observed, these quantities replace the corresponding DA
assumptions and define the updated operating condition for the
remaining horizon.

Given the observed state and the price guidance $\mathbf{u}_{\tau}$
provided by the Pricing Agent, the RT optimizer computes a revised
schedule as in Eq.~\eqref{eq:rt_objective}:
\begin{equation}
\mathbf{s}_{\tau}^{\mathrm{RT},*}
\in
\arg\min_{\mathbf{s}\in\mathcal{S}_{\tau}^{\mathrm{RT}}(\mathbf{u}_{\tau},\mathbf{x}_{\tau}^{\mathrm{obs}})}
C_{\tau:T}^{\mathrm{PTO}}\!\left(\mathbf{u}_{\tau},\mathbf{s}\right),
\label{eq:rt_objective}
\end{equation}
where $\mathbf{s}_{\tau}^{\mathrm{RT},*}$ is the revised RT schedule,
$\mathcal{S}_{\tau}^{\mathrm{RT}}(\cdot)$ is the feasible set updated
with the observed RT conditions, and $C_{\tau:T}^{\mathrm{PTO}}(\cdot)$
is the PTO energy-trading cost over the remaining horizon.
The RT model preserves the same physical constraints as the DA model;
the difference is that selected inputs are updated with observations
before the model is rerun.
The main RT updates are:
\begin{align}
e_{k,\tau} &= e_{k,\tau}^{\mathrm{obs}},
&& \forall k\in\mathcal{K},
\label{eq:rt_initial_soc}\\
\xi_i^{\mathrm{RT}} &= \xi_i^{\mathrm{obs}},
&& \forall i\in\mathcal{I},
\label{eq:rt_energy_consumption}\\
\pi_t^g &= \pi_t^{g,\mathrm{obs}},
&& \forall t\in\mathcal{T}_{\tau}^{\mathrm{RT}},
\label{eq:rt_price_update}\\
b_{k,i,t} &\ge \bar{b}_{k,i,t}^{\mathrm{active}},
&& \forall k\in\mathcal{K},\forall i\in\mathcal{I},\forall t\in\mathcal{T}_{\tau}^{\mathrm{RT}}.
\label{eq:rt_active_trip}
\end{align}
Constraint~\eqref{eq:rt_initial_soc} sets the current battery energy
of each bus equal to the observed value at the moment of
re-optimization.
Constraint~\eqref{eq:rt_energy_consumption} updates the energy
requirement of each route using the observed route-level consumption
value.
Constraint~\eqref{eq:rt_price_update} replaces the DA
electricity-price profile with the updated price profile used from the
current interval onward.
Constraint~\eqref{eq:rt_active_trip} preserves trips that are already
underway, preventing the RT optimizer from interrupting active service
commitments.

If no re-optimization is needed, the system continues executing the
last accepted reference schedule $\mathbf{s}_{\mathrm{ref}}$.
Once a revised schedule is accepted by the Evaluator Agent, it becomes
the new reference $\mathbf{s}_{\tau}^{\mathrm{RT},*}$ for subsequent
RT updates.

\subsection{Agentic Supervisory Flow}
\label{sec:agentic_logic}

This section provides the formal mathematical definitions of the agent
policies and the connection between agent-generated price guidance and
the PTO scheduling model.
In a classical aggregator design
(e.g.,~\cite{manzolli2024hierarchical}), a dedicated upper-level
problem would maximize aggregator revenue $\Pi^{\mathrm{Agg}}$ by
directly optimizing the tariff vectors $\boldsymbol{\rho}^{+}$ and
$\boldsymbol{\rho}^{-}$ offered to the PTO:
\begin{equation}
\Pi^{\mathrm{Agg}}
=
\sum_{t \in \mathcal{T}}
\left[
\rho_{p(t)}^{+} w_t^{+}\Delta t
-
\rho_{p(t)}^{-} w_t^{-}\Delta t
-
\pi_t^{g}\!\left(w_t^{+}-w_t^{-}\right)\Delta t
\right].
\label{eq:agg_revenue}
\end{equation}
In the proposed framework, this optimization problem is replaced by the
agentic supervisory layer:
\begin{equation}
\underbrace{
\max_{\boldsymbol{\rho}^{+},\boldsymbol{\rho}^{-}}
\Pi^{\mathrm{Agg}}
}_{\text{classical aggregator upper level}}
\quad
\Longrightarrow
\quad
\underbrace{
\left(
\pi_{\theta_T}^{T},
\pi_{\theta_P}^{P},
\pi_{\theta_E}^{E}
\right)
}_{\text{agentic supervisory layer}},
\label{eq:upper_replacement}
\end{equation}
where $\pi_{\theta_T}^{T}$, $\pi_{\theta_P}^{P}$, and
$\pi_{\theta_E}^{E}$ denote the Trigger Agent, Pricing Agent, and Evaluator Agent policies, respectively. The aggregator function is therefore redistributed across these specialized agents, which determine when the optimizer should be called, which price guidance should be tested, and whether the resulting schedule should be accepted. The Pricing Agent maps the current decision context and coordination mode to a price-guidance vector:
\begin{equation}
\mathbf{u}_{\tau}
=
\pi_{\theta_P}^{P}\left(\mathbf{c}_{\tau},m\right),
\label{eq:pricing_policy}
\end{equation}
where the decision context $\mathbf{c}_{\tau}$ is:
\begin{equation}
    \mathbf{c}_{\tau} = \Psi\!\left(\mathbf{f}_{\tau},\,
    \mathbf{g}_{\tau},\,\mathbf{h}_{\tau},\,m\right),
    \label{eq:context}
\end{equation}
with $\mathbf{f}_{\tau}$ the fleet-state vector, $\mathbf{g}_{\tau}$
the grid and market vector, $\mathbf{h}_{\tau}$ the historical memory
vector, and $m$ the coordination mode.
The guidance vector $\mathbf{u}_{\tau}$ is converted into the tariff
vectors fed to the PTO scheduling model as:
\begin{equation}
\rho_p^{+} = \alpha_p^{+}\bar{\pi}_{p}^{g},
\qquad
\rho_p^{-} = \alpha_p^{-}\bar{\pi}_{p}^{g},
\qquad
\forall p\in\mathcal{P},
\label{eq:agentic_tariffs}
\end{equation}
where $\alpha_p^{+}$ and $\alpha_p^{-}$ are the buy and sell multipliers proposed by the Pricing Agent, and $\bar{\pi}_{p}^{g}$ is the average grid price in tariff period $p$. In the DA workflow, given the price guidance $\mathbf{u}_{\tau}$ from Eq.~\eqref{eq:pricing_policy}, the Evaluator Agent assesses the resulting schedule:
\begin{equation}
    \left(a_{\tau},\, r_{\tau},\, \gamma_{\tau}\right)
    = \pi^{E}_{\theta_{E}}\!\left(\mathbf{c}_{\tau},\, \mathbf{u}_{\tau}\right),
    \label{eq:da_policies}
\end{equation}
where $a_{\tau} \in \{\texttt{accept},\texttt{rerun}\}$ is the
evaluation decision, $r_{\tau}$ is the supervisory rationale, and
$\gamma_{\tau} \in [0,1]$ is a confidence score.
In the RT workflow, the Trigger Agent policy is evaluated first:
\begin{equation}
    \delta_{\tau} = \pi^{T}_{\theta_{T}}\!\left(\mathbf{c}_{\tau},\,
    \mathbf{s}^{*}_{\tau-1}\right),
    \label{eq:trigger_policy}
\end{equation}
where $\delta_{\tau} \in \{\texttt{skip},\texttt{optimize}\}$ is the
trigger decision and $\mathbf{s}^{*}_{\tau-1}$ is the last accepted
plan.
If $\delta_{\tau} = \texttt{optimize}$, the Pricing Agent and Evaluator
Agent are invoked as in Eqs.~\eqref{eq:pricing_policy}
and~\eqref{eq:da_policies}.
In both workflows, the optimization engine solves:
\begin{equation}
    \mathbf{s}_{\tau}^{*}
    \in \arg\min_{\mathbf{s}\,\in\,\mathcal{S}(\mathbf{u}_{\tau})}
    C^{\mathrm{PTO}}\!\left(\mathbf{u}_{\tau},\,\mathbf{s}\right),
    \label{eq:guided_lower}
\end{equation}
connecting the agentic price guidance directly to the feasible
scheduling decision defined in Eq.~\eqref{eq:lower}.
In the RT case this specializes to Eq.~\eqref{eq:rt_objective}, where
the feasible set expands to
$\mathcal{S}_{\tau}^{\mathrm{RT}}(\mathbf{u}_{\tau},
\mathbf{x}_{\tau}^{\mathrm{obs}})$ to incorporate the observed state.

\setcounter{table}{0}
\setcounter{figure}{0}
\setcounter{equation}{0}
\renewcommand{\thetable}{B.\arabic{table}}
\renewcommand{\thefigure}{B.\arabic{figure}}
\renewcommand{\theequation}{B.\arabic{equation}}

\section{Compact Prompt Templates}
\label{app:compact_prompt_templates}
\setcounter{table}{0}
\setcounter{figure}{0}
\setcounter{equation}{0}
\renewcommand{\thetable}{\thesection.\arabic{table}}
\renewcommand{\thefigure}{\thesection.\arabic{figure}}
\renewcommand{\theequation}{\thesection.\arabic{equation}}

This appendix reports compact versions of the prompt templates used in the agent experiments. The original prompts contain implementation placeholders, repeated formatting instructions, and full multiplier/output-schema constraints; the templates below preserve the operational content needed to reproduce the experimental logic without reproducing the complete workflow text. The placeholders \texttt{\{mode\}}, \texttt{\{spot\_prices\}}, \texttt{\{fleet\_context\}}, and \texttt{\{optimization\_result\}} denote values injected by the workflow at runtime.

\begin{table*}[t]
\centering
\caption{Prompt paradigms used for the DA Pricing Agent experiments.}
\label{tab:app_prompt_paradigms}
\footnotesize
\begin{adjustbox}{max width=\textwidth}
\begin{tabular}{lllL{8.2cm}}
\toprule
Acronym & Name & Agent & Compact definition \\
\midrule
ZS & Zero-shot & DA Pricing Agent & Role, economic framework, mode objective, hard constraints, and structured output requirements are provided, but no worked examples or explicit reasoning scaffold is included. \\
CoT & Chain-of-thought & DA Pricing Agent & The zero-shot prompt is augmented with an explicit reasoning scaffold that asks the agent to reason about price levels, V2G opportunity, tariff spread, and the selected profit-based or operational-based objective before returning the final multiplier vectors. \\
FS & Few-shot & DA Pricing Agent & The zero-shot prompt is augmented with compact examples that illustrate mode-consistent tariff behavior for profit-based and operational-based pricing. \\
FS+CoT & Few-shot + chain-of-thought & DA Pricing Agent & The prompt combines examples with the reasoning scaffold, giving the agent both behavioral demonstrations and a structured path for translating the objective into buy/sell multipliers. \\
RT-ZS-S & Zero-shot with constrained schema & RT Trigger, Pricing, and Evaluator Agents & RT agents use role prompts and structured context without DA prompt-paradigm variants; outputs are constrained to workflow fields such as action, trigger type, confidence, accepted update, and multiplier vectors. \\
\bottomrule
\end{tabular}
\end{adjustbox}
\end{table*}

\medskip
\noindent\textit{DA Pricing Agent zero-shot template.}
\begin{quote}
\footnotesize
You are the Electric Bus Aggregator (EBA) price-setting agent for DA optimization. Set buy and sell multipliers for 48 half-hour timesteps. The EBA buys from the grid at spot price \(P_t\), sells charging energy to the PTO at \(m^{buy}_tP_t\), buys V2G energy from the PTO at \(m^{sell}_tP_t\), and resells V2G energy to the grid at \(P_t\). Aggregator profit is the sum of charging margin and V2G margin. If \texttt{\{mode\}} is profit-based, maximize aggregator revenue while preserving feasible fleet operation and some V2G opportunity. If \texttt{\{mode\}} is operational-based, minimize PTO cost while preserving positive aggregator revenue and feasible operation. Respect hard constraints: return exactly 48 buy multipliers and 48 sell multipliers; keep \(m^{buy}_t>1\), \(m^{sell}_t<1\); use the price forecast \texttt{\{spot\_prices\}} and fleet context \texttt{\{fleet\_context\}}; output only the requested structured fields.
\end{quote}

\medskip
\noindent\textit{Chain-of-thought extension.}
\begin{quote}
\footnotesize
Before producing the final multiplier vectors, reason through the coupled tariff decision: identify low- and high-price periods; determine when charging should be encouraged or discouraged; determine when V2G export is valuable; check whether buy multipliers are too high to create surplus SOC; check whether sell multipliers are too low to induce V2G; then adapt the multipliers to the selected profit-based or operational-based objective. Return only the final structured multiplier output, not the reasoning text.
\end{quote}

\medskip
\noindent\textit{Few-shot extension.}
\begin{quote}
\footnotesize
Use the examples as behavioral guidance. A profit-based example raises buy multipliers and lowers sell multipliers enough to increase aggregator margin, but avoids eliminating V2G volume. An operational-based example keeps buy multipliers close to the spot price and sell multipliers high enough to transfer more V2G value to the PTO, while preserving small positive aggregator revenue. Generalize the pattern to the provided price profile and fleet context rather than copying the example values.
\end{quote}

\medskip
\noindent\textit{Few-shot plus chain-of-thought template.}
\begin{quote}
\footnotesize
Combine the few-shot behavioral examples with the reasoning scaffold. First use the examples to identify the intended mode behavior; then reason about price periods, charging incentives, V2G incentives, margin, PTO exposure, and feasibility; finally return exactly 48 buy and sell multipliers in the required structured format. The examples define the behavioral direction, while the reasoning scaffold adapts that direction to the current price and fleet context.
\end{quote}

\medskip
\noindent\textit{DA Evaluator Agent compact template.}
\begin{quote}
\footnotesize
Evaluate the DA optimization result \texttt{\{optimization\_result\}} under the selected mode. Inspect PTO daily cost, aggregator revenue, bought energy, sold energy, feasibility status, terminal SOC, and whether the result improves on the best known solution. In profit-based mode, accept results that improve or preserve aggregator revenue without infeasibility or excessive loss of V2G opportunity. In operational-based mode, accept results that reduce PTO cost while preserving feasibility and non-negative aggregator revenue. If the result should be revised, return adjusted guidance with exactly 48 buy and 48 sell multipliers; otherwise return an accept decision and rationale in the required structured format.
\end{quote}

\medskip
\noindent\textit{RT zero-shot constrained-output template.}
\begin{quote}
\footnotesize
At each RT timestep, read the current system state, active reference plan, price deviation, energy deviation, delay status, previous trigger history, and remaining horizon. The Trigger Agent returns a schema-constrained decision: skip or optimize, trigger type, confidence, flagged buses, and rationale. If optimization is triggered, the Pricing Agent returns updated buy/sell guidance for the remaining horizon using the current mode objective. The Evaluator Agent then accepts, rejects, or requests a bounded rerun based on feasibility, PTO cost, aggregator revenue, V2G activity, flagged-bus deviations, and terminal SOC. These RT prompts are zero-shot role prompts with constrained output fields rather than the ZS/CoT/FS/FS+CoT DA prompt variants.
\end{quote}

\bibliographystyle{elsarticle-num}
\bibliography{references}

\end{document}